\documentclass[10pt,twocolumn,letterpaper]{article}

\usepackage{cvpr}
\usepackage{times}
\usepackage{epsfig}
\usepackage{graphicx}
\usepackage{amsmath}
\usepackage{amssymb}
\usepackage{mynotation}
\graphicspath{{./figures/}}
\usepackage{booktabs}
\usepackage[caption=false]{subfig}
\usepackage[pagebackref=true,breaklinks=true,letterpaper=true,colorlinks,bookmarks=false,hypertexnames=false]{hyperref}

\cvprfinalcopy 

\def\cvprPaperID{2354} 

\begin{document}
	
	\title{Density Adaptive Point Set Registration}
	
\author{Felix J\"aremo Lawin, Martin Danelljan, Fahad Shahbaz Khan, Per-Erik Forss\'en, Michael Felsberg \\
	\small Computer Vision Laboratory, Department of Electrical Engineering, Link\"oping University, Sweden\\
	\small\{\texttt{felix.jaremo-lawin},\; \texttt{martin.danelljan},\; \texttt{fahad.khan},\; \texttt{per-erik.forssen},\; \texttt{michael.felsberg}\}\texttt{@liu.se}
	{}
}
	
	\maketitle
	\thispagestyle{empty}
	
\begin{abstract}
	Probabilistic methods for point set registration have demonstrated competitive results in recent years. These techniques estimate a probability distribution model of the point clouds. While such a representation has shown promise, it is highly sensitive to variations in the density of 3D points. This fundamental problem is primarily caused by changes in the sensor location across point sets.	
	We revisit the foundations of the probabilistic registration paradigm. Contrary to previous works, we model the underlying structure of the scene as a latent probability distribution, and thereby induce invariance to point set density changes. Both the probabilistic model of the scene and the registration parameters are inferred by minimizing the Kullback-Leibler divergence in an Expectation Maximization based framework. Our density-adaptive registration successfully handles severe density variations commonly encountered in terrestrial Lidar applications. We perform extensive experiments on several challenging real-world Lidar datasets. The results demonstrate that our approach outperforms state-of-the-art probabilistic methods for multi-view registration, without the need of re-sampling. Code is available at \href{https://github.com/felja633/DARE}{https://github.com/felja633/DARE}.
\end{abstract}

\section{Introduction}

\begin{figure}
	\centering
    \includegraphics[height=32mm]{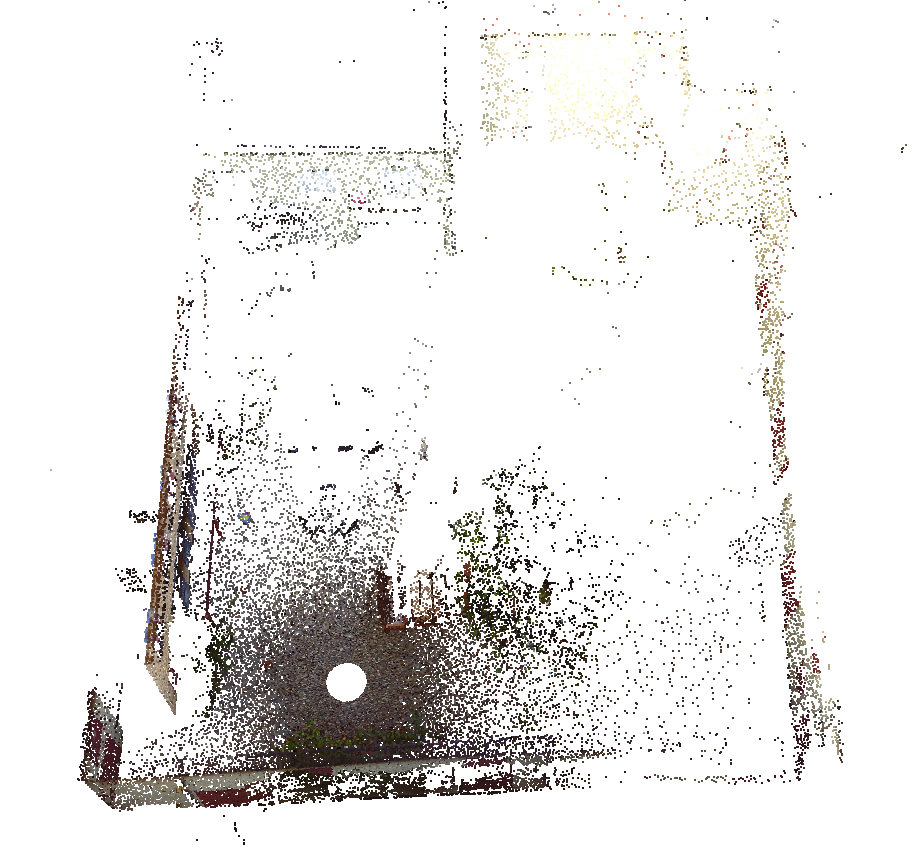}\hspace{7mm}%
	\includegraphics[height=32mm]{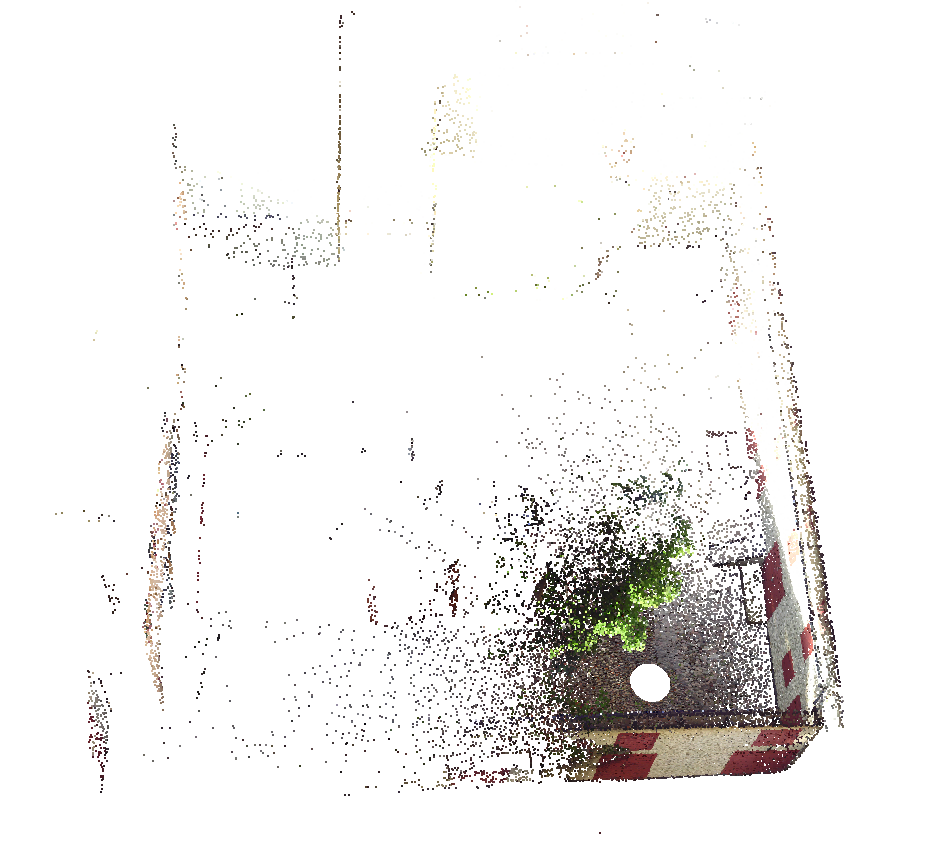}
	\includegraphics[trim=0 0 0 120,clip,height=31mm]{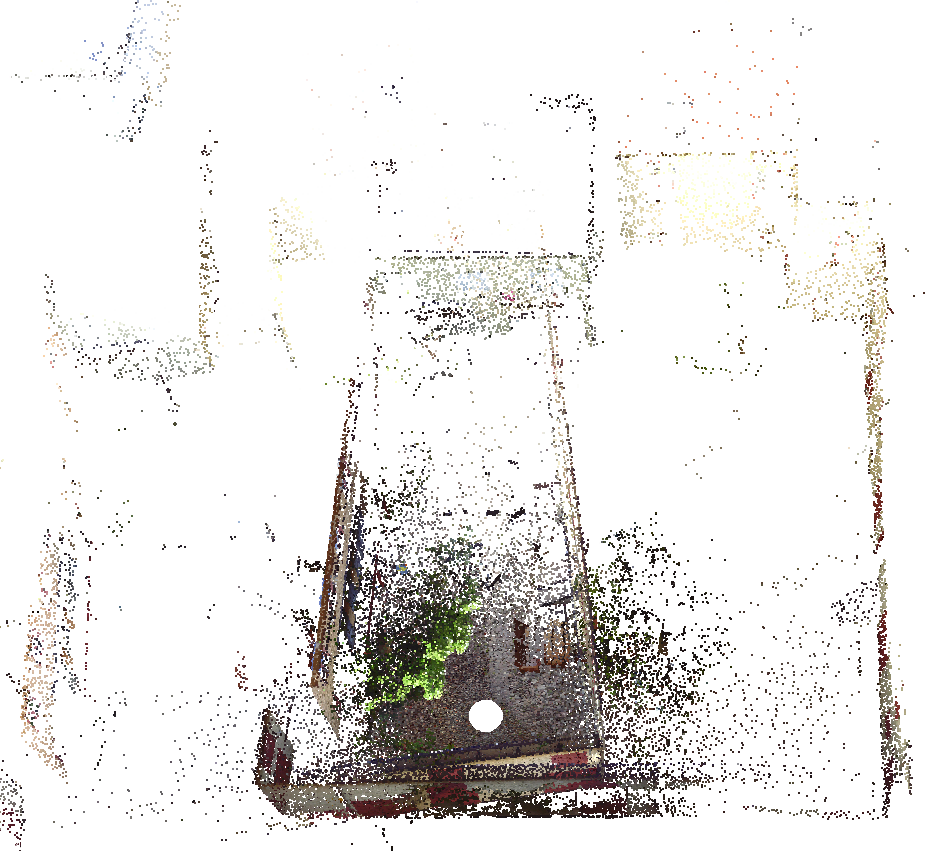}\hspace{6mm}%
	\includegraphics*[trim=100 0 130 80,clip,height=31mm]{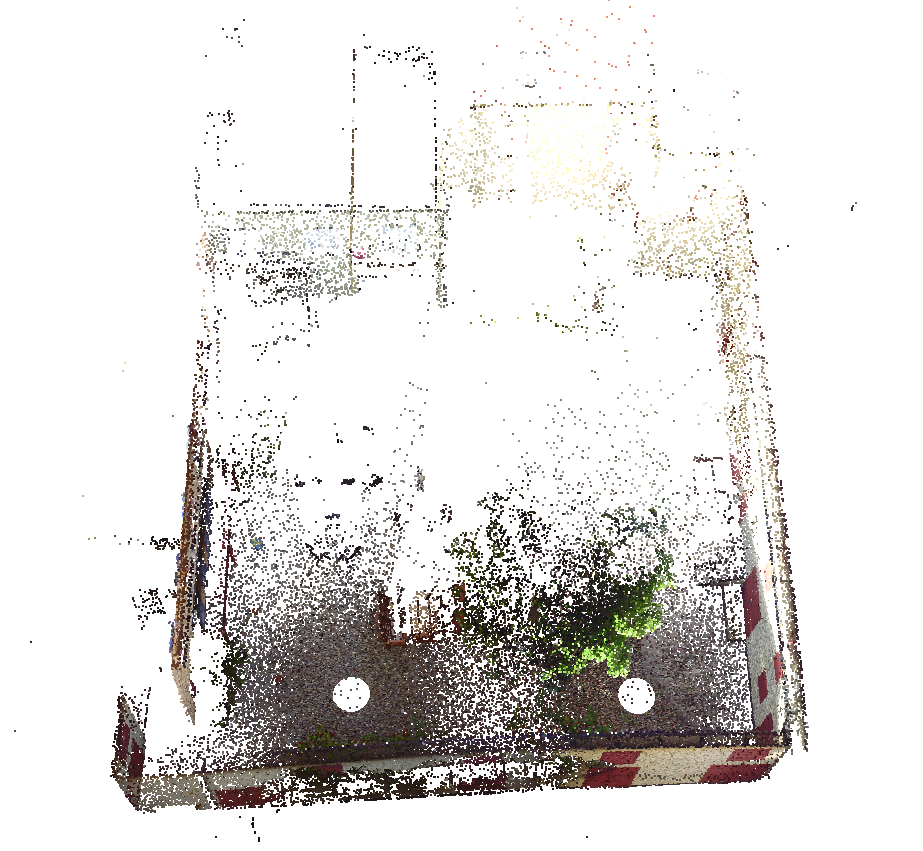}
	\includegraphics*[trim=0 100 0 30,clip,height=32mm]{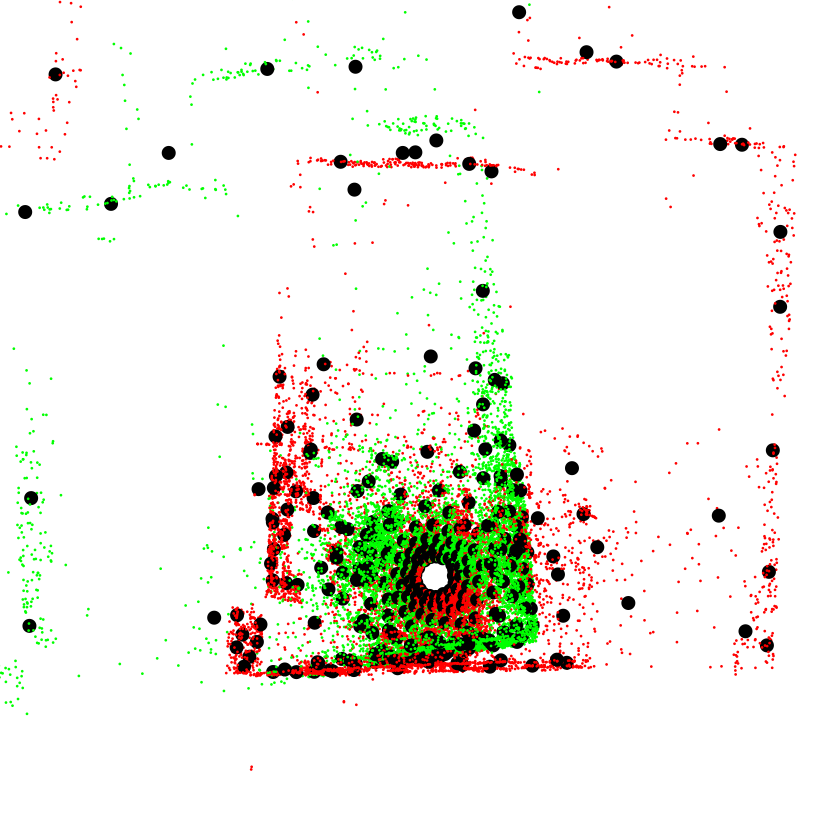}\hspace{7mm}%
	\includegraphics*[trim=100 10 0 88,clip,height=31mm]{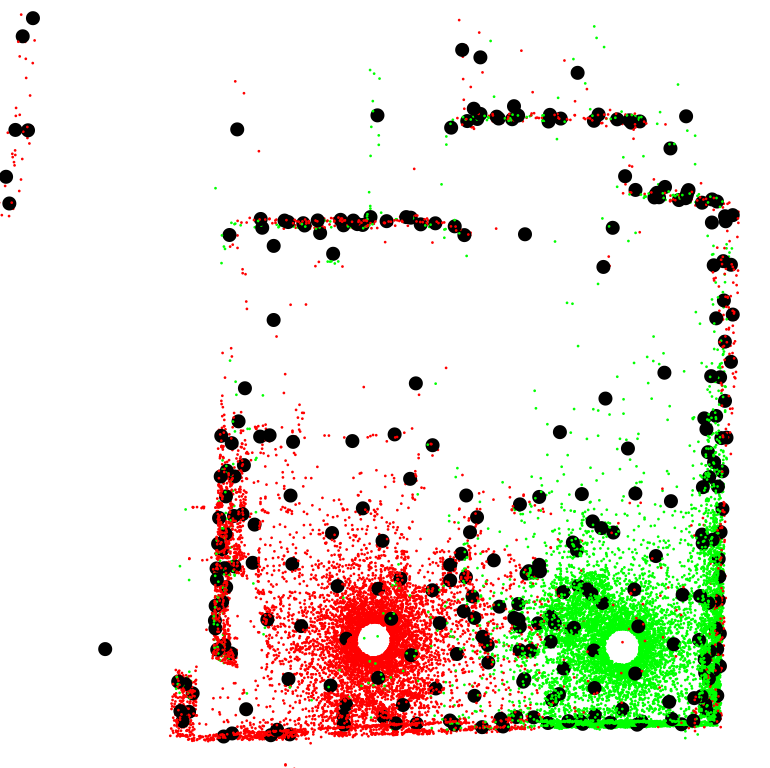}%
	\caption{Two example Lidar scans (top row), with significantly varying density of 3D-points. State-of-the-art probabilistic method \cite{DanelljanCVPR2016} (middle left) only aligns the regions with high density. This is caused by the emphasis on dense regions, as visualized by the Gaussian components in the model (black circles in bottom left). Our method (right) successfully exploits essential information available in sparse regions, resulting in accurate registration.}\vspace{-2mm}
	\label{fig:intro}
\end{figure}

3D-point set registration is a fundamental problem in computer vision, with applications in 3D mapping and scene understanding. Generally, the point sets are acquired using a 3D sensor, \eg a Lidar or an RGBD camera. The task is then to align point sets acquired at different positions, by estimating their relative transformations. Recently, probabilistic registration methods have shown competitive performance in different scenarios, including pairwise \cite{myronenko2010point,HoraudPAMI11,GMMregPAMI11} and multi-view registration \cite{evangelidis2014generative,DanelljanCVPR2016}.

In this work, we revisit the foundations of the probabilistic registration paradigm, leading to a reformulation of the Expectation Maximization (EM) based approaches \cite{evangelidis2014generative, DanelljanCVPR2016}. In these approaches, a Maximum Likelihood (ML) formulation is used to simultaneously infer the transformation parameters, and a Gaussian mixture model (GMM) of the point distribution. Our formulation instead minimizes the Kullback-Leibler divergence between the mixture model and a latent scene distribution.

Common acquisition sensors, including Lidar and RGBD cameras, do not sample all surfaces in the scene with a uniform density (figure~\ref{fig:intro}, top row). The density of 3D-point observations is highly dependent on (1) the distance to the sensor, (2) the direction of the surface relative to the sensor, and (3) inherent surface properties, such as specularity. Despite recent advances, state-of-the art probabilistic methods \cite{myronenko2010point,evangelidis2014generative,DanelljanCVPR2016,GMMregPAMI11,HoraudPAMI11} struggle 
under varying sampling densities, in particular when the translational part of the transformation is significant.

The density variation is problematic for standard ML-based approaches since each 3D-point corresponds to an observation with equal weight.
Thus, the registration focuses on regions with high point densities, while neglecting sparse regions. 

This negligence is clearly visible in figure~\ref{fig:intro} (bottom left), where registration has been done using CPPSR \cite{DanelljanCVPR2016}. Here the vast majority of Gaussian components (black circles) are located in regions with high point densities.
A common consequence of this is inaccurate or failed registrations.
Figure~\ref{fig:intro} (middle right) shows an example registration using our approach. Unlike the existing method \cite{DanelljanCVPR2016}, our model exploits information available in both dense and sparse regions of the scene, as shown by the distribution of Gaussian components (figure~\ref{fig:intro}, bottom right).

\subsection{Contributions}
We propose a probabilistic point set registration approach that counters the issues induced by sampling density variations. Our approach directly models the underlying structure of the 3D scene using a novel density-adaptive formulation. The probabilistic scene model and the transformation parameters are jointly inferred by minimizing the Kullback-Leibler (KL) divergence with respect to the latent scene distribution. This is enabled by modeling the acquisition process itself, explicitly taking the density variations into account. To this end, we investigate two alternative strategies for estimating the acquisition density: a model-based and a direct empirical method. Experiments are performed on several challenging Lidar datasets, demonstrating the effectiveness of our approach in difficult scenarios with drastic variations in the sampling density.

\section{Related work}
The problem of 3D-point set registration is extensively pursued in computer vision. Registration methods can be coarsely categorized into local and global methods. Local methods rely on an initial estimate of the relative transformation, which is then iteratively refined. The typical example of a local method is the Iterative Closest Point (ICP) algorithm. In ICP, registration is performed by iteratively alternating between establishing point correspondences and refining the relative transformation. While the standard ICP \cite{ICP_PAMI92} benefits from a low computational cost, it is limited by a narrow region of convergence. Several works \cite{segal2009generalized,Softassign97,TriICP_IVC05} investigate how to improve the robustness of ICP.

Global methods instead aim at finding the global solution to the registration problem. Many global methods rely on local ICP-based or probabilistic methods and use, \eg, multiple restarts \cite{ICP_restarts}, graph optimization \cite{theiler2015globally}, branch-and-bound \cite{Campbell_2016_CVPR} techniques to search for a globally optimal registration. Another line of research is to use feature descriptors to find point correspondences in a robust estimation framework, such as RANSAC \cite{raguram2008comparative}. Zhou \etal \cite{zhou2016fast} also use feature correspondences, but minimize a Geman-McClure robust loss. A drawback of such global methods is the reliance on accurate geometric feature extraction.

Probabilistic registration methods model the distribution of points as a density function. These methods perform alignment either by employing a correlation based approach or using an EM based optimization framework. In correlation based approaches \cite{tsin2004correlation,GMMregPAMI11}, the point sets are first modeled separately as density functions. The relative transformation between the points set is then obtained by minimizing a metric or divergence between the densities. These methods lead to nonlinear optimization problems with non-convex constraints. Unlike correlation based methods, the EM based approaches \cite{myronenko2010point,evangelidis2014generative} find an ML-estimate of the density model and transformation parameters.

Most methods implicitly assume a uniform density of the point clouds, which is hardly the case in most applications. 
The standard approach \cite{rusu20113d} to alleviate the problems of varying point density is to re-sample the point clouds in a separate preprocessing step. The aim of this strategy is to achieve an approximately uniform distribution of 3D points in the scene. A common method is to construct a voxel grid and taking the mean point in each voxel. Comparable uniformity is achieved using the Farthest Point Strategy \cite{eldar1997farthest}, were points are selected iteratively to maximize the distance to neighbors. Geometrically Stable Sampling (GSS) \cite{gelfand2003geometrically} also incorporates surface normals in the sample selection process. However, such re-sampling methods have several shortcomings. First, 3D scene information is discarded as observations are grouped together or removed, leading to sparsification of the point cloud. Second, the sampling rate, \eg voxel size, needs to be hand picked for each scenario as it depends on the geometry and scale of the point cloud. Third, a suitable trade-off between uniformity and sparsity must be found. Thus, such preprocessing steps are complicated and their efficacy is questionable. In this paper, we instead explicitly model the density variations induced by the sensor.

There exist probabilistic registration methods that tackle the problem of non-uniform sampling density \cite{campbell2015adaptive,hermans2011robust}. In \cite{campbell2015adaptive}, a one class support vector machine is trained for predicting the underlying density of partly occluded point sets. The point sets are then registered by minimizing the L2 distance between the density models. In \cite{latecki2006new}, an extended EM framework for modeling noisy data points is derived, based on minimizing the KL divergence. This framework was later exploited for outlier handling in point set registration \cite{hermans2011robust}. 
Unlike these methods, we introduce a latent distribution of the scene and explicitly model the point sampling density using either a sensor model or an empirical method. 

\section{Method}

In this work, we revisit probabilistic point cloud registration, with the aim of alleviating the problem of non-uniform point density. To show the impact of our model, we employ the Joint Registration of Multiple Point Clouds (JRMPC) \cite{evangelidis2014generative}. Compared to previous probabilistic methods, JRMPC has the advantage of enabling joint registration of multiple input point clouds. Furthermore, this framework was recently extended to use color \cite{DanelljanCVPR2016}, geometric feature descriptors \cite{DanelljanICPR2016} and incremental joint registration \cite{evangelidis2017}. However, our approach can be applied to a variety of other probabilistic registration approaches. Next, we present an overview of the baseline JRMPC method.

\subsection{Probabilistic Point Set Registration}
\label{sec:pcreg}
Point set registration is the problem of finding the relative geometric transformations between $M$ different sets of points. We directly consider the general case where $M \geq 2$. Each set $\mathcal{X}_i = \{x_{ij}\}_{j=1}^{N_i}, i = 1,\ldots, M$, consists of 3D-point observations $x_{ij} \in \reals^3$ obtained from, e.g., a Lidar scanner or an RGBD camera. We let capital letters $X_{ij}$ denote the associated random variables for each observation. In general, probabilistic methods aim to model the probability densities $p_{X_i}(x)$, for each point set $i$, using for instance Gaussian Mixture Models (GMMs).

Different from previous approaches, JRMPC derives the densities $p_{X_i}(x)$ from a global probability density model $p_V(v|\theta)$, which is defined in a reference coordinate frame given parameters $\theta$. The registration problem can then be formulated as finding the relative transformations from point set $\mathcal{X}_i$ to the reference frame. We let $\phi(\cdot; \omega) : \reals^3 \rightarrow \reals^3$ be a 3D transformation parametrized by $\omega \in \reals^D$. The goal is then to find the parameters $\omega_i$ of the transformation from $\mathcal{X}_i$ to the reference frame, such that $\phi(X_{ij}; \omega_i) \sim p_V$. Similarly to previous works \cite{evangelidis2014generative,DanelljanCVPR2016}, we focus on the most common case of rigid transformation $\phi(x;\omega) = R_\omega x + t_\omega$. In this case, the density model of each point set is obtained as $p_{X_i}(x|\omega_i,\theta) = p_V(\phi(x;\omega_i)|\theta)$.

The density $p_V(v|\theta)$ is composed by a mixture of Gaussian distributions,
\begin{equation}
\label{eq:GMM}
p_V(v|\theta) = \sum_{k=1}^K \pi_k \norm (v;\mu_k, \Sigma_k) \,.
\end{equation}
Here, $\norm (v;\mu, \Sigma)$ is a Gaussian density with mean $\mu$ and covariance $\Sigma$. The number of components is denoted by $K$ and $\pi_k$ is the prior weight of component $k$. The set of all mixture parameters is thus $\theta = \{\pi_k, \mu_k, \Sigma_k\}_{k=1}^K$.

Different from previous works, the mixture model parameters $\theta$ and transformation parameters $\omega$ are inferred jointly in the JRMPC framework, assuming independent observations. This is achieved by maximizing the log-likelihood function,
\begin{equation}
\label{eq:ML}
\mathcal{L}(\Theta;\mathcal{X}_1,\ldots,\mathcal{X}_M) = \sum_{i}^M \sum_j^{N_i} \log (p_V(\phi(x_{ij};\omega_i)|\theta))\,.
\end{equation}
Here, we denote the set of all parameters in the model as $\Theta = \{\theta, \omega_1, \ldots, \omega_M\}$. Inference is performed with the Expectation Maximization (EM) algorithm, by first introducing a latent variable $Z \in \{1, \ldots, K\}$ that assigns a 3D-point $V$ to a particular mixture component $Z=k$. The complete data likelihood is then given by $p_{V,Z}(v,k|\theta) = p_Z(k|\theta) p_{V|Z}(v|k,\theta)$, where $p_Z(k|\theta) = \pi_k$ and $p_{V|Z}(v|k,\theta) = \norm (v;\mu_k, \Sigma_k)$. The original mixture model \eqref{eq:GMM} is recovered by marginalizing the complete data likelihood over the latent variable $Z$.

The E-step in the EM algorithm involves computing the expected complete-data log likelihood,
\begin{equation}
\label{eq:CDL}
Q(\Theta;\!\Theta^n)\!=\!\!\sum_{i}^M \sum_j^{N_i} E_{Z|x_{ij},\Theta^n}\!\!\left[ \log (p_{V,Z}(\phi(x_{ij};\omega_i),Z|\theta)) \right].
\end{equation}
Here, the conditional expectation is taken over the latent variable given the observed point $x_{ij}$ and the current estimate of the model parameters $\Theta^n$. In the M-step, the model parameters are updated as $\Theta^{n+1} = \argmax_\Theta Q(\Theta; \Theta^n)$. This process is then repeated until convergence.

\subsection{Sampling Density Adaptive Model}
To tackle the issues caused by non-uniform point densities, we revise the underlying formulation and model assumptions. Instead of modeling the density of 3D-points, we aim to infer a model of the actual 3D-structure of the scene. To this end, we introduce the latent probability distribution of the scene $q_V(v)$. Loosely defined, it is seen as a uniform distribution on the observed surfaces in the scene. Intuitively, $q_V(v)$ encodes all 3D-structure, i.e.\ walls, ground, objects etc., that is measured by the sensor. Different models of $q_V(v)$ are discussed is section~\ref{sec:sampling-func}. Technically, $q_V$ might not be absolutely continuous and is thus regarded a probability measure. However, we will denote it as a density function to simplify the presentation.

Our goal is to model $q_V(v)$ as a parametrized density function $p_V(v|\theta)$. We employ a GMM \eqref{eq:GMM} and minimize the Kullback-Leibler (KL) divergence from $p_V$ to $q_V$,
\begin{equation}
\text{KL}(q_V||p_V) = \int \!\log\left(\frac{q_{V}(v)}{p_V(v|\theta)}\right) q_V(v) \,\diff v \,.
\label{eq:KL}
\end{equation}
Utilizing the decomposition of the KL-divergence $\text{KL}(q_V||p_V) =  H(q_V, p_V) - H(q_V)$ into the cross entropy $H(q_V, p_V)$ and entropy $H(q_V)$ of $q_V$, we can equivalently maximize,
\begin{equation}
\label{eq:objective}
\mathcal{E}(\Theta) = -H(q_V, p_V) = \int \!\log\left(p_V(v|\theta)\right) q_V(v) \diff v
\end{equation}

In \eqref{eq:objective}, the integration is performed in the reference frame of the scene. On the other hand, the 3D points $x_{ij}$ are observed in the coordinate frames of the individual sensors. As in section~\ref{sec:pcreg}, we relate these coordinate frames with the transformations $\phi(\cdot; \omega_i)$. By applying the change of variables $v = \phi(x; \omega_i)$, we obtain
\begin{align}
\label{eq:objective_x}
\mathcal{E}(\Theta) = \frac{1}{M} \sum_{i=1}^{M} \int_{\reals^3} & \!\log\left(p_V(\phi(x; \omega_i)|\theta)\right) \cdot \\
&q_V(\phi(x; \omega_i)) |\!\det(D\phi(x;\omega_i))| \,\diff x \,. \nonumber
\end{align}
Here, $|\!\det(D\phi(x;\omega_i))|$ is the determinant of the Jacobian of the transformation. From now on, we assume rigid transformations, which implies $|\!\det(D\phi(x;\omega_i))| = 1$.

We note that if $\{x_{ij}\}_{i=1}^{N_i}$ are independent samples from $q_V(\phi(x; \omega_i))$, the original maximum likelihood formulation \eqref{eq:ML} is recovered as a Monte Carlo sampling of the objective \eqref{eq:objective_x}. Therefore, the conventional ML formulation \eqref{eq:ML} relies on the assumption that the observed points $x_{ij}$ follow the underlying uniform distribution of the scene $q_V$. However, this assumption completely neglects the effects of the acquisition sensor. Next, we address this problem by explicitly modeling the sampling process.

In our formulation, we consider the points in set $i$ to be independent samples $x_{ij} \sim q_{X_i}$ of a distribution $q_{X_i}(x)$. In addition to the 3D structure $q_V$ of the scene, $q_{X_i}$ can also depend on the position and properties of the sensor, and the inherent properties of the observed surfaces. This enables more realistic models of the sampling process to be employed. By assuming that the distribution $q_V$ is absolutely continuous \cite{durrett2010probability} w.r.t.\ $q_{X_i}$, eq.\ \eqref{eq:objective_x} can be written,
\begin{equation}
\label{eq:objective_q}
\mathcal{E}(\Theta) \!= \!\sum_{i=1}^{M}\! \int_{\reals^3} \!\!\!\!\log\left(p_V(\phi(x; \omega_i)|\theta)\right) \!\frac{q_V(\phi(x; \omega_i))}{q_{X_i}(x)} q_{X_i}(x) \,\diff x .
\end{equation}
Here, we have also ignored the factor $1/M$. The fraction $f_i(x) = \frac{q_V(\phi(x; \omega_i))}{q_{X_i}(x)}$ is known as the Radon-Nikodym derivative \cite{durrett2010probability} of the probability distribution $q_V(\phi(x; \omega_i))$ with respect to $q_{X_i}(x)$. 

Intuitively, $f_i(x)$ is the ratio between the density in the latent scene distribution and the density of points in point cloud $\mathcal{X}_i$. Since it weights the observed 3D-points based on the local density, we term it the \emph{observation weighting function}. In section~\ref{sec:sampling-func}, we later introduce two different approximations of $f_i(x)$ to model the sampling process itself.

\subsection{Inference}
\label{sec:inference}

In this section, we describe the inference algorithm used to minimize \eqref{eq:objective_q}. We show that the EM-based framework used in \cite{evangelidis2014generative,DanelljanCVPR2016} also generalizes to our model. As in section~\ref{sec:pcreg}, we apply the latent variable $Z$ and the complete-data likelihood $p_{V,Z}(v,k|\theta)$. We define the expected complete-data cross entropy as,
\begin{align}
	\label{eq:EM1}
	&\mathcal{Q}(\Theta,\Theta^n) = \\
	&\sum_{i=1}^{M}\! \int_{\reals^3} \!\!\!E_{Z|x,\Theta^n}\! \left[\log\left(p_{V,Z}(\phi(x; \omega_i),Z|\theta)\right) \right] f_i(x) q_{X_i}(x) \,\diff x . \nonumber
\end{align}
Here, $\Theta^n$ is the current estimate of the parameters. 
The E-step involves evaluating the expectation in \eqref{eq:EM1}, taken over the probability distribution of the latent variable,
\begin{align}
	\label{eq:latent_posterior}
	p_{Z|X_i}(k|x,\Theta) &= \frac{p_{X_i,Z}(x,k|\Theta)}{\sum_{k=1}^{K} p_{X_i,Z}(x,k|\Theta)} \nonumber\\
	& = \frac{\pi_k \norm (\phi(x;\omega_k); \mu_k, \Sigma_k)}{\sum_{l=1}^K \pi_l \norm (\phi(x;\omega_l); \mu_l, \Sigma_l)} \,.
\end{align}

To maximize \eqref{eq:EM1} in the M-step, we first perform a Monte Carlo sampling of \eqref{eq:EM1}. Here we use the assumption that the observations are independent samples drawn from $x_{ij} \sim q_{X_i}$. To simplify notation, we define $\alpha_{ijk}^n = p_{Z|X_i}(k|x_{ij},\Theta^n)$. Then \eqref{eq:EM1} is approximated as,
\begin{align}
	\label{eq:EM2}
	\mathcal{Q}&(\Theta,\Theta^n) \approx Q(\Theta,\Theta^n) = \\
	&\sum_{i=1}^{M} \frac{1}{N_i} \sum_{j=1}^{N_i} \sum_{k=1}^{K} \alpha_{ijk}^n f_i(x_{ij}) \log\left(p_{V,Z}(\phi(x_{ij}; \omega_i),k|\theta)\right) \,. \nonumber
\end{align}
Please refer to the supplementary material for a detailed derivation of the EM procedure.

The key difference of \eqref{eq:EM2} compared to the ML case \eqref{eq:CDL}, is the weight factor $f_i(x_{ij})$. This factor effectively weights each observation $x_{ij}$ based on the local density of 3D points. Since the M-step has a form similar to \eqref{eq:CDL}, we can apply the optimization procedure proposed in \cite{evangelidis2014generative}. Specifically, we employ two conditional maximization steps \cite{meng1993}, to optimize over the mixture parameters $\theta$ and transformation parameters $\omega_i$ respectively. Furthermore, our approach can be extended to incorporate color information using the approach proposed in \cite{DanelljanCVPR2016}.

\subsection{Observation Weights}
\label{sec:sampling-func}
We present two approaches of modeling the observation weight function $f_i(x)$. The first is based on a sensor model, while the second is an empirical estimation of the density.

\subsubsection{Sensor Model Based}
\label{sec:sensor_model}
Here, we estimate the sampling distribution $q_{X_i}$ by modeling the acquisition sensor itself. For this method we therefore assume that the type of sensor (e.g.\ Lidar) is known and that each point set $\mathcal{X}_i$ consists of a single scan. The latent scene distribution $q_V$ is modeled as a uniform distribution on the observed surfaces $S$. That is, $S$ is a 2-dimensional manifold consisting of all observable surfaces. Thus, we define $q_V(A) = \frac{1}{|S|}\int_{S \cap A} \diff S$ for any measurable set $A \subset \reals^3$. For simplicity, we use the same notation $q_V(A) = \mathbb{P}(V \in A)$ for the probability measure $q_V$ of $V$. We use $|S| = \int_{S} \diff S$ to denote the total area of $S$. 

We model the sampling distribution $q_{X_i}$ based on the properties of a terrestrial Lidar. It can however be extended to other sensor geometries, such as time-of-flight cameras. We can without loss of generality assume that the Lidar is positioned in the origin $x=0$ of the sensor-based reference frame in $\mathcal{X}_i$. Further, let $S_i = \phi_i^{-1}(S)$ be the scene transformed to the reference frame of the sensor. Here, we use $\phi_i(x) = \phi(x,\omega_i)$ to simplify notation. We note that the density of Lidar rays is decreasing quadratically with distance. For this purpose, we model the Lidar as light source emitting uniformly in all directions of its field of view. The sampling probability density at a visible point $x \in S_i$ is then proportional to the absorbed intensity, calculated as $\frac{\hat{n}_x\tp \hat{x}}{\|x\|^2}$. Here, $\hat{n}_x$ is the unit normal vector of $S_i$ at $x$, $\|\cdot\|$ is the Euclidean norm and $\hat{x} = x/\|x\|$.

The sampling distribution is defined as the probability of observing a point in a subset $A \subset \reals^3$. It is obtained by integrating the point density over the part of the surface $S$ intersecting $A$,
\begin{equation}
	\label{eq:sensor_model}
	q_{X_i}(A) \!=\! \int_{S_i\cap A} \hspace{-1mm} \frac{g_i}{|S|} \, \diff S_i , \hspace{2mm} g_i(x) \!=\! \begin{cases} a \frac{\hat{n}_x\tp \hat{x}}{\|x\|^2} \hspace{-3mm}&, x \in S_i \cap F_i \\ \varepsilon &, \text{otherwise} \end{cases}
\end{equation}
Here, $F_i \subset \reals^3$ is the observed subset of the scene, $\varepsilon$ is the outlier density and $a$ is a constant such that the probability integrates to 1. Using the properties of $q_V$, we can rewrite \eqref{eq:sensor_model} as $q_{X_i}(A) = \int_A g_i \,\diff (q_V \circ \phi_i)$. Here, $q_V \circ \phi_i$ is the composed measure $q_V(\phi_i(A))$. From the properties of the Radon-Nikodym derivative \cite{durrett2010probability}, we obtain that $f_i = \frac{\diff (q_V \circ \phi_i)}{\diff q_{X_i}} = \frac{1}{g_i}$. In practice, surface normal estimates can be noisy, thus promoting the use of a regularized quotient $f_i(x) = a \frac{\|x\|^2}{\gamma \hat{n}_x\tp \hat{x} + 1 - \gamma}$, for some fix parameter $\gamma \in [0,1]$. Note that the calculation of $f_i(x)$ only requires information about the distance $\|x\|$ to the sensor and the normal $\hat{n}_x$ of the point cloud at $x$. For details and derivations, see the supplementary material.

\begin{figure}
	\centering
	\includegraphics*[trim=0 0 0 0,width=0.48\columnwidth]{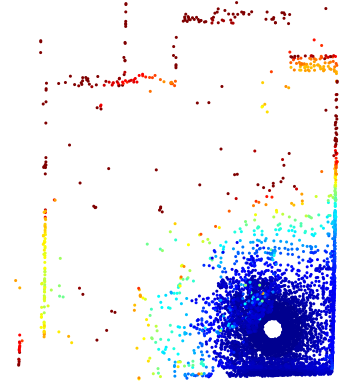}\hspace{1mm}%
	\includegraphics*[trim=0 0 0 0,width=0.48\columnwidth]{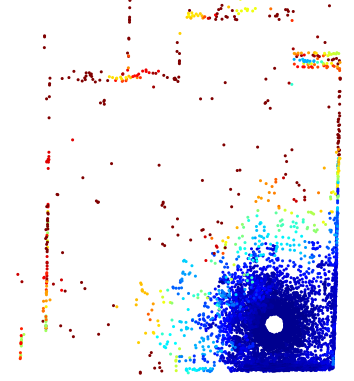}
	\includegraphics*[trim=0 2 0 0,width=0.4\columnwidth]{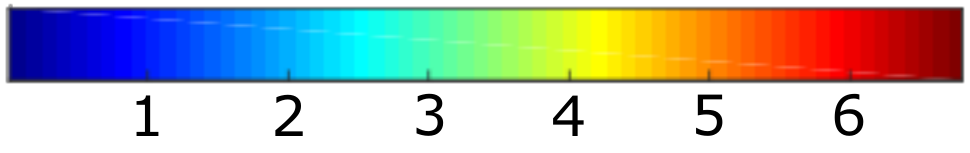}\vspace{-1mm}
	\caption{Visualization of the observation weight computed using our sensor based model (left) and empirical method (right). The 3D-points in the densely sampled regions in the vicinity of the Lidar are assigned low weights, while the impact of points in the sparser regions are increased. The two approaches produce visually similar results. The main differences are seen in the transitions from dense to sparser regions.}\vspace{-2mm}
	\label{fig:observation_weights}
\end{figure}

\subsubsection{Empirical Sample Model}
\label{sec:empirical_model}
As an alternative approach, we propose an empirical model of the sampling density. Unlike the sensor-based model in section~\ref{sec:sensor_model}, our empirical approach does not require any information about the sensor. It can thus be applied to arbitrary point clouds, without any prior knowledge. We modify the latent scene model $q_V$ from sec.~\ref{sec:sensor_model} to include a 1-dimensional Gaussian distribution in the normal direction of the surface $S$. This uncertainty in the normal direction models the coarseness or evenness of the surface, which leads to variations orthogonal to the underlying surface. In the local neighborhood of a point $\bar{v} \in S$, we can then approximate the latent scene distribution as a 1-dimensional Gaussian in the normal direction $q_V(v) \approx \frac{1}{|S|} \norm ( \hat{n}_{\bar{v}}\tp(v - \bar{v}); 0, \sigma_{\hat{n}}^2(\bar{v}))$. It is motivated by a locally planar approximation of the surface $S$ at $\bar{v}$, where $q_V(v)$ is constant in the tangent directions of $S$. Here, $\sigma_{\hat{n}}^2(\bar{v})$ is the variance in the normal direction.

To estimate the observation weight function $f(x) = \frac{q_V(\phi(x))}{q_X(x)}$, we also find a local approximation of the sampling density $q_X(x)$. For simplicity, we drop the point set index $i$ in this section and assume a rigid transformation $\phi(x) = Rx+t$. First, we extract the $L$ nearest neighbors $x_1, \ldots, x_L$ of the 3D point $x$ in the point cloud. We then find the local mean $\bar{x} = \frac{1}{L} \sum_l x_l$ and covariance $C = \frac{1}{L-1} \sum_l (x_l-\bar{x})\tp (x_l-\bar{x})$. This yields the local sampling density estimate $q_X(x) \approx \frac{L}{N} \norm (x; \bar{x}, C)$. Let $C = B D B\tp$ be the eigenvalue decomposition of $C$ with $B = (\hat{b}_1, \hat{b}_2, \hat{b}_3)$ and $D = \text{diag} (\sigma_1^2, \sigma_2^2, \sigma_3^2)$, and eigenvalues sorted in descending order. Since we assume the points to originate from a locally planar region, we deduce that $\sigma_1^2, \sigma_2^2 \gg \sigma_3^2$. Furthermore, $\hat{b}_3$ and $\sigma_3^2$ approximate the normal direction of the surface and the variance in this direction. We utilize this information for estimating the local latent scene distribution, by setting $\bar{v} = \phi(\bar{x})$, $\hat{n}_{\bar{v}} = R \hat{b}_3$ and $\sigma^2_{\hat{n}}(\bar{v}) = \sigma_3^2$. We then obtain,
\begin{equation}
	\label{eq:empirical_weights}
	f(x) = \frac{q_V(\phi(x))}{q_X(x)} \propto \sigma_1 \sigma_2 e{\raisebox{7pt}{\scriptsize$\frac{1}{2} (x - \bar{x})\tp B 
		{\tiny\arraycolsep=0.2\arraycolsep\ensuremath{ \begin{pmatrix}
		\sigma_1^{-2}&0&0 \\ 0&\sigma_2^{-2}&0 \\ 0&0&0
		\end{pmatrix}}}
	B\tp (x - \bar{x})$}} .
\end{equation}
Here, we have omitted proportionality constants independent of the point location $x$ in $f(x)$, since they do not influence the objective \eqref{eq:objective_q}.
A detailed derivation is provided in the supplementary material. In practice, we found $f(x) \propto \sigma_1 \sigma_2$ to be a sufficiently good approximation since $\sigma_1^{-2}, \sigma_2^{-2} \approx 0$ and $\bar{x} \approx x$. 

Note that the observation weights $f_i(x_{ij})$ in \eqref{eq:EM2} can be precomputed once for every registration. The added computational cost of the density adaptive registration method is therefore minimal and in our experiments we only observed an increase in computational time of $2 \%$   compared to JRMPC. In figure~\ref{fig:observation_weights}, the observation weights $f_i(x_{ij})$ are visualized for both the sensor based model (left) and empirical method (right).

\section{Experiments}
We integrate our sampling density adaptive model in the probabilistic framework JRMPC \cite{evangelidis2014generative}. Furthermore, we evaluate our approach, when using feature information, by integrating the model in the color based probabilistic method CPPSR \cite{DanelljanCVPR2016}. 

First we perform a synthetic experiment to highlight the impact of sampling density variations on point set registration. Second, we perform quantitative and qualitative evaluations on two challenging Lidar scan datasets: Virtual Photo Sets \cite{VPSoutdoor} and the ETH TLS \cite{theiler2015globally}. Further detailed results are presented in the supplementary material. 

\subsection{Experimental Details}
\label{sec:exp_det}
Throughout the experiments we randomly generate ground-truth rotations and translations for all point sets. The point sets are initially transformed using this ground-truth. The resulting point sets are then used as input for all compared registration methods. For efficiency reasons we construct a random subset of 10k points for each scan in all the datasets. The experiments on the point sets from VPS and ETH TLS are conducted in two settings. First, we perform direct registration on the constructed point sets. Second, we evaluate all compared registration methods, except for our density adaptive model, on re-sampled point sets. The registration methods without density adaptation, however, are sensitive to the choice of re-sampling technique and sampling rate. In the supplementary material we provide an exhaustive evaluation of FPS \cite{eldar1997farthest}, GSS \cite{gelfand2003geometrically} and voxel grid re-sampling at different sampling rates. We then extract the best performing re-sampling settings for each registration method and use it in the comparison as an empirical upper bound in performance.

\noindent\textbf{Method naming:} We evaluate two main variants of the density adaptive model. In the subsequent performance plots and tables, we denote our approach using the sensor model based observation weights in section \ref{sec:sensor_model} by DARS, and the empirical observation weights in section \ref{sec:empirical_model} by DARE.
 
\noindent\textbf{Parameter settings:} We use the same values for all the parameters that are shared between our methods and the two baselines: the JRMPC and CPPSR. As in \cite{evangelidis2014generative}, we use a uniform mixture component to model the outliers. In our experiments, we set the outlier ratio $0.005$ and fix the spatial component weights $\pi_k$ to uniform. In case of pairwise registration, we set the number of spatial components $K=200$. In the joint registration scenario, we set $K=300$ for all methods to increase the capacity of the model for larger scenes. We use 50 EM iterations for both the pairwise and joint registration scenarios. In case of color features, we use 64 components as proposed in \cite{DanelljanCVPR2016}.

In addition to the above mentioned parameters, we use the $L=10$ nearest neighbors to estimate $\sigma_1$ and $\sigma_2$ in section~\ref{sec:empirical_model}. To regularize the observation weights $f_i(x_{ij})$ (section \ref{sec:sampling-func}) and remove outlier values, we first perform a median filtering using the same neighborhood size of $L=10$ points. We then clip all the observation weights that exceed a certain threshold. We fix this threshold to 8 times the mean value of all observation weights within a point set. In the supplementary material we provide an analysis of these parameters and found our method not to be sensitive to the parameter values. For the sensor model approach (section~\ref{sec:sensor_model}) we set $\gamma = 0.9$. We keep all parameters fix in all experiments and datasets. 

\noindent\textbf{Evaluation Criteria:} The evaluation is performed by computing the
angular error (\ie the {\it geodesic distance}) between the found rotation, $R$, and the ground-truth rotation, $R_\text{gt}$. This distance is computed via the Frobenius distance $d_F(R,R_\text{gt})$, using the relation $d_G(R_1,R_2)=2\sin^{-1}(d_F(R_1,R_2)/\sqrt{8})$, which is derived in \cite{hartley13}. 
To evaluate the performance in terms of robustness, we report the failure rate as the percentage of registrations with an angular error greater than 4 degrees. Further, we present the accuracy in terms of the mean angular error among inlier registrations. In the supplementary material we also provide the translation error.

\begin{figure}
	\centering
		\includegraphics[width=0.69\columnwidth]{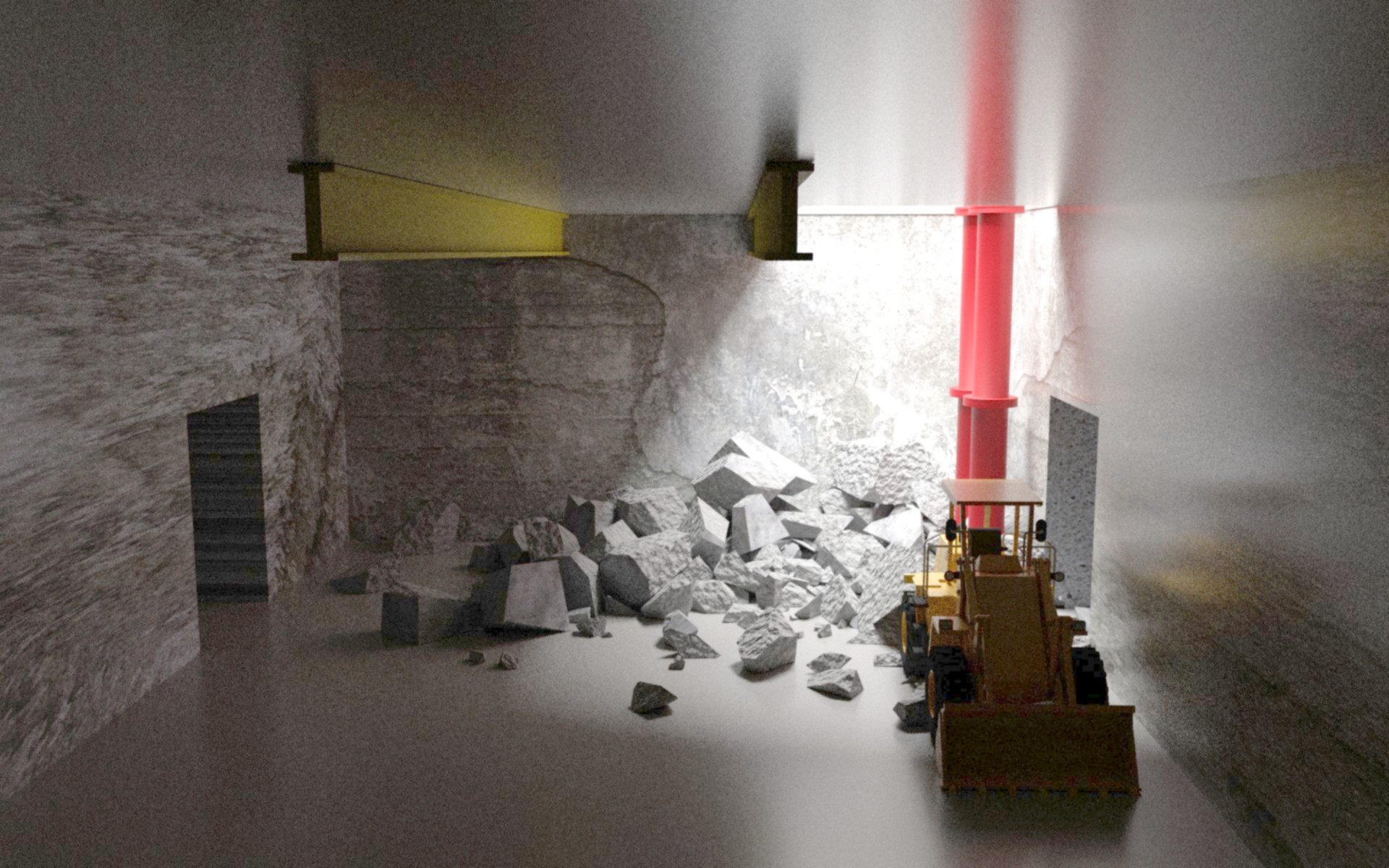}
		\includegraphics[width=0.3\columnwidth]{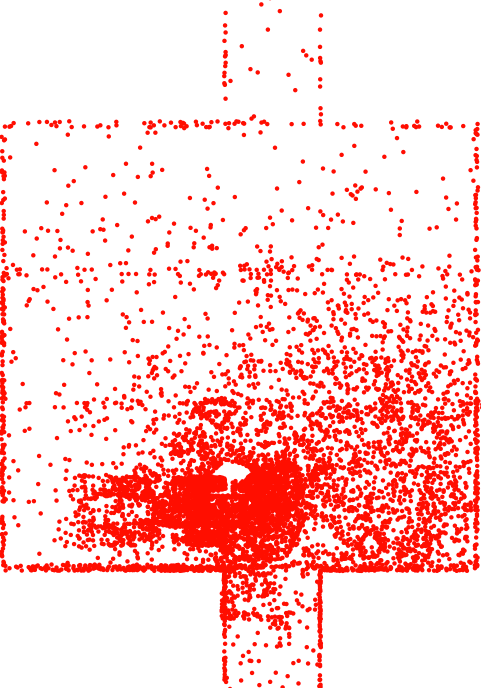}
	\caption{The synthetic 3D scene. Left: Rendering of the scene. Right: Top view of re-sampled point set with varying density.}
	\label{fig:synt}
\end{figure}

\subsection{Synthetic Data}
We first validate our approach on a synthetic dataset to isolate the impact of sampling density variations on pairwise registration. 
We construct synthetic point clouds by performing point sampling on a polygon mesh that simulates an indoor 3D scene (see figure \ref{fig:synt} left). We first sample uniformly, and densely. We then randomly select a virtual sensor location. Finally, we simulate Lidar sampling density variations by randomly removing points according to their distances to the sensor position (see figure \ref{fig:synt} right). In total the synthetic dataset contains 500 point set pairs.

\begin{figure}
  \centering\vspace{-3mm}
		\subfloat[Synthetic\label{fig:synthetic_recall}]{\includegraphics[width=0.49\columnwidth]{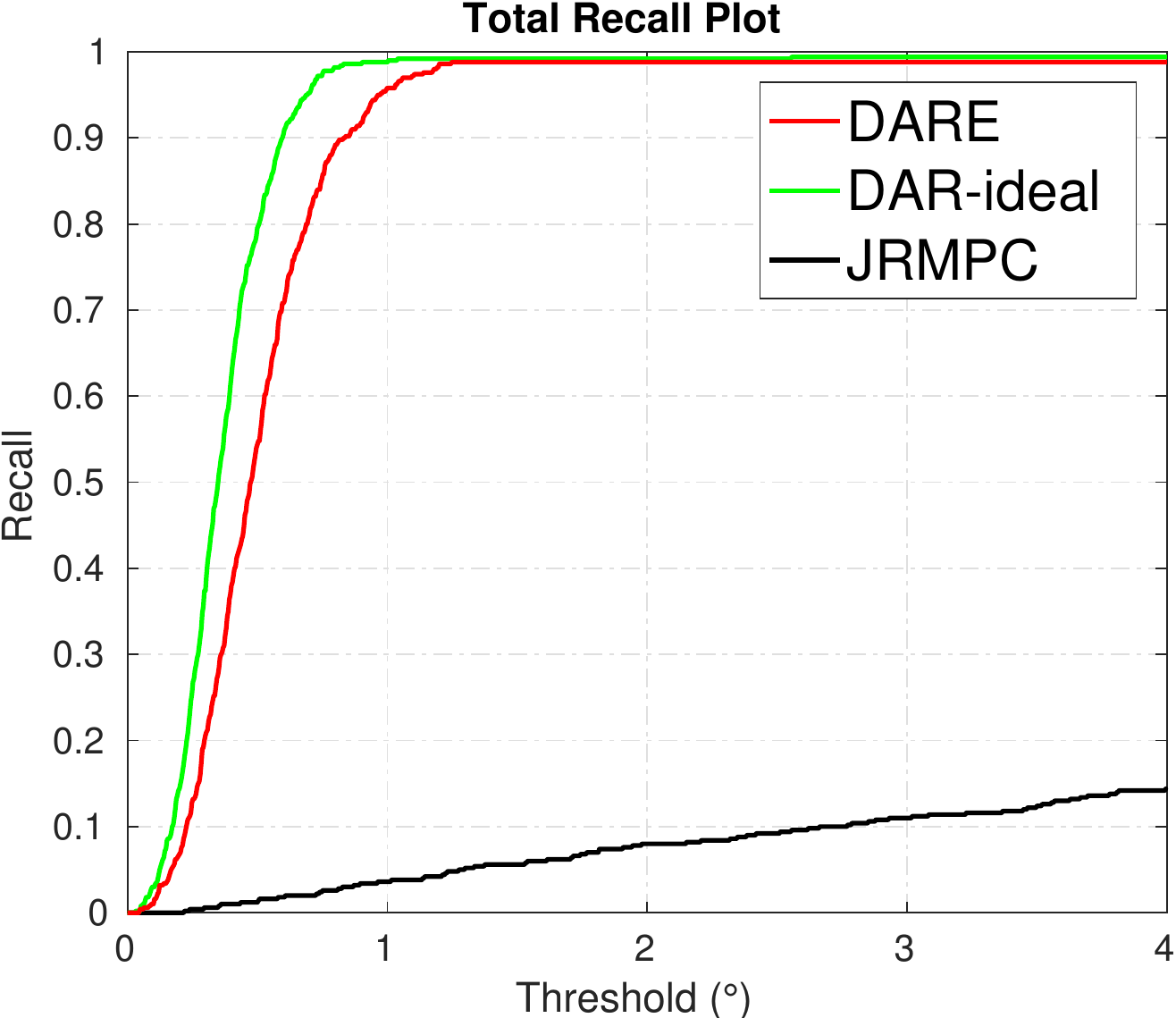}}
		\subfloat[Combined VPS and TLS\label{fig:joint_vpstls}]{\includegraphics[width=0.49\columnwidth]{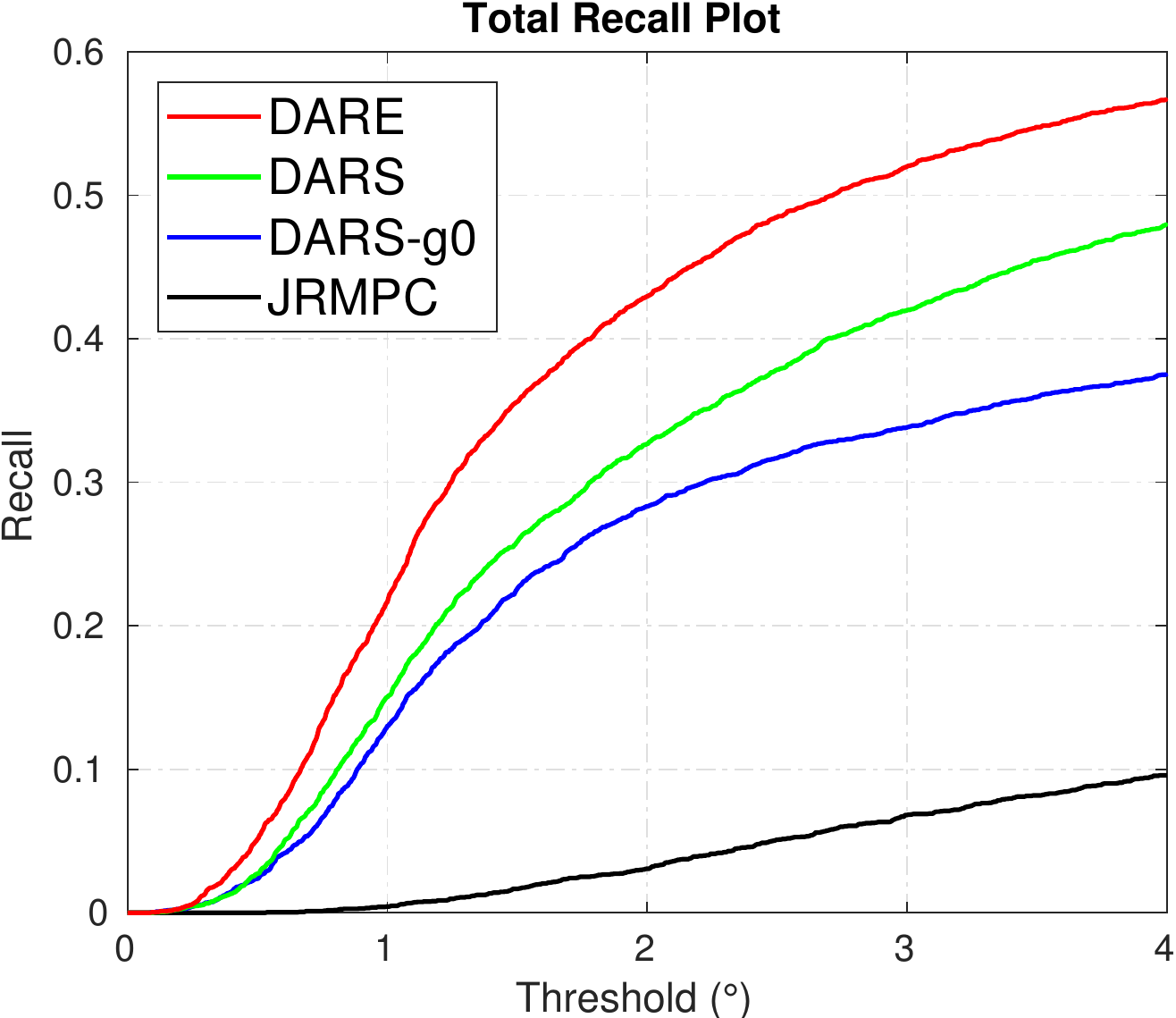}}
\caption{Recall curves with respect to the angular error. (a) Results on the synthetic dataset. Our DARE approach closely follows the upper bound, DAR-ideal. (b) Results on the combined VPS and TLS ETH datasets. In all cases, our DARE approach significantly improves over the baseline JRMPC \cite{evangelidis2014generative}.}\vspace{-2mm}
	\label{fig:synthetic}
\end{figure}

Figure~\ref{fig:synthetic_recall} shows the recall curves, plotting the ratio of registrations with an angular error smaller than a threshold. We report results for the baseline JRMPC and our DARE method. We also report the results when using the ideal sensor sample model to compute the observation weights $f_i(x_{ij})$, called DAR-ideal. Note that the same sampling function was employed in the construction of the virtual scans. This method therefore corresponds to an upper performance bound of our DARE approach.

The baseline JRMPC model struggles in the presence of sampling density variations, providing inferior registration results with a failure rate of 85 $\%$. Note that the JRMPC corresponds to setting the observation weights to uniform $f_i(x_{ij})=1$ in our approach. The proposed DARE, significantly improves the registration results by reducing the failure rate from 85 $\%$ to 2 $\%$. Further, the registration performance of DARE closely follows the ideal sampling density model, demonstrating the ability of our approach to adapt to sampling density variations. 

\begin{table}
	\centering
	\resizebox{0.8\columnwidth}{!}{%
		\begin{tabular}{lcc}
			\toprule
			&Avg. inlier error ($^\circ$)&Failure rate (\%) \\\midrule
			JRMPC&2.44$\pm$0.87&90.4\\
			JRMPC\emph{-eub}&1.67$\pm$0.92&46.0\\
			ICP&1.73$\pm$1.04&62.6\\
			ICP\emph{-eub}&1.81$\pm$0.99&55.7\\
			CPD&1.88$\pm$1.25&90.0\\
			CPD\emph{-eub}&1.30$\pm$0.95&40.8\\\midrule
			\textbf{DARE}&1.45$\pm$0.89&43.3\\\bottomrule
		\end{tabular}
	}\vspace{1mm}
	\caption{A comparison of our approach with existing methods in terms of average inlier angular error and failure rate for pairwise registration on the combined VPS and TLS ETH dataset. The methods with the additional \emph{-eub} in the name are the empirical upper bounds using re-sampling. Our DARE method improves over the baseline JRMPC, regardless of re-sampling settings, both in terms of accuracy and robustness.}\vspace{-2mm}
	\label{tab:full}
\end{table}

\subsection{Pairwise Registration}
\label{sec:pairwise}
We perform pairwise registration experiments on the joint Virtual Photo Set (VPS) \cite{VPSoutdoor} and the TLS ETH \cite{theiler2015globally} datasets. The VPS dataset consists of Lidar scans from two separate scenes, each containing four scans. The TLS ETH dataset consists of two separate scenes, with seven and five scans respectively. We randomly select pairs of different scans within each scene, resulting in total 3720 point set pairs. The ground-truth for each pair is generated by first randomly selecting a rotation axis. We then rotate one of the point sets with a rotation angle (within 0-90 degrees) around the rotation axis and apply a random translation, drawn from a multivariate normal distribution with standard deviation 1.0 meters in all directions. 

\begin{figure*}[t!]
	\centering\vspace{-5mm}%
	\subfloat[CPPSR\label{fig:CPPSR}]{\includegraphics*[trim=100 150 100 160,width=0.4\textwidth]{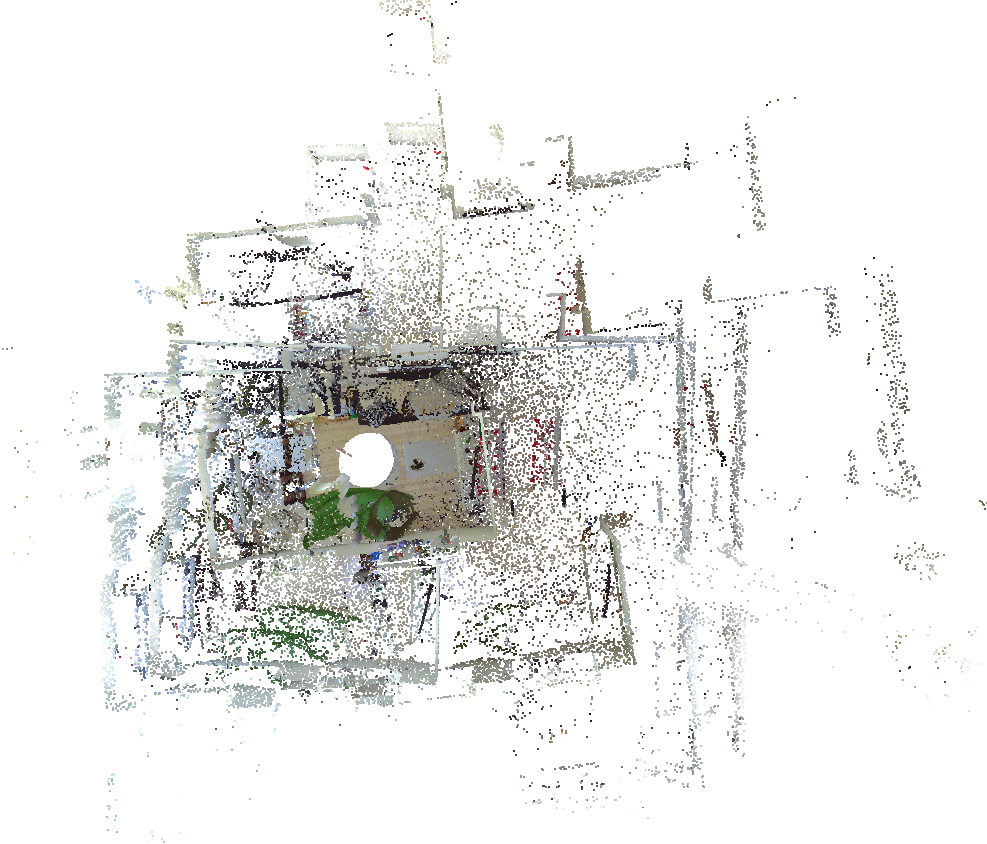}}\hspace{10mm}%
	\subfloat[DARE-color\label{fig:DAREcolor}]{\includegraphics*[trim=70 200 70 30,width=0.45\textwidth]{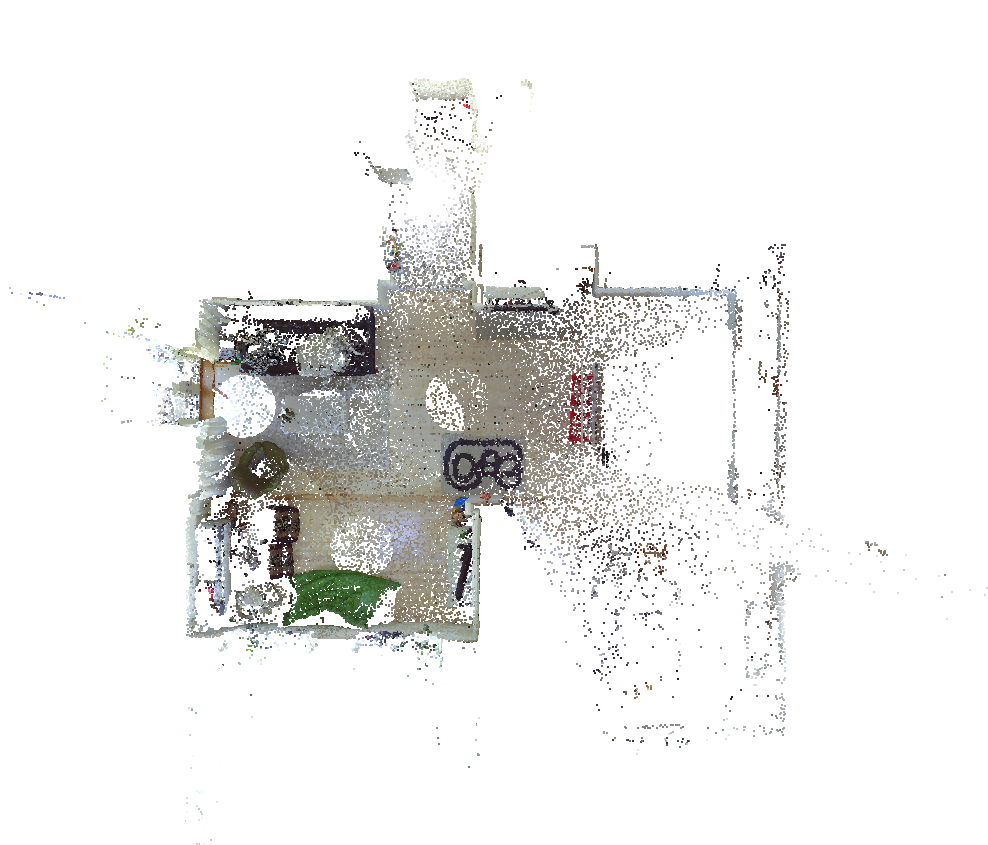}}
	\vspace{1mm}
	\caption{Joint registration of the four point sets in the VPS indoor dataset. (a) CPPSR \cite{DanelljanCVPR2016} only aligns the high density regions and neglects sparsely sampled 3D-structure. (b) Corresponding registration using our density adaptive model incorporating color information.}
	\label{fig:multi}
\end{figure*}

\begin{figure}
	\centering
	\includegraphics[width=0.8\columnwidth]{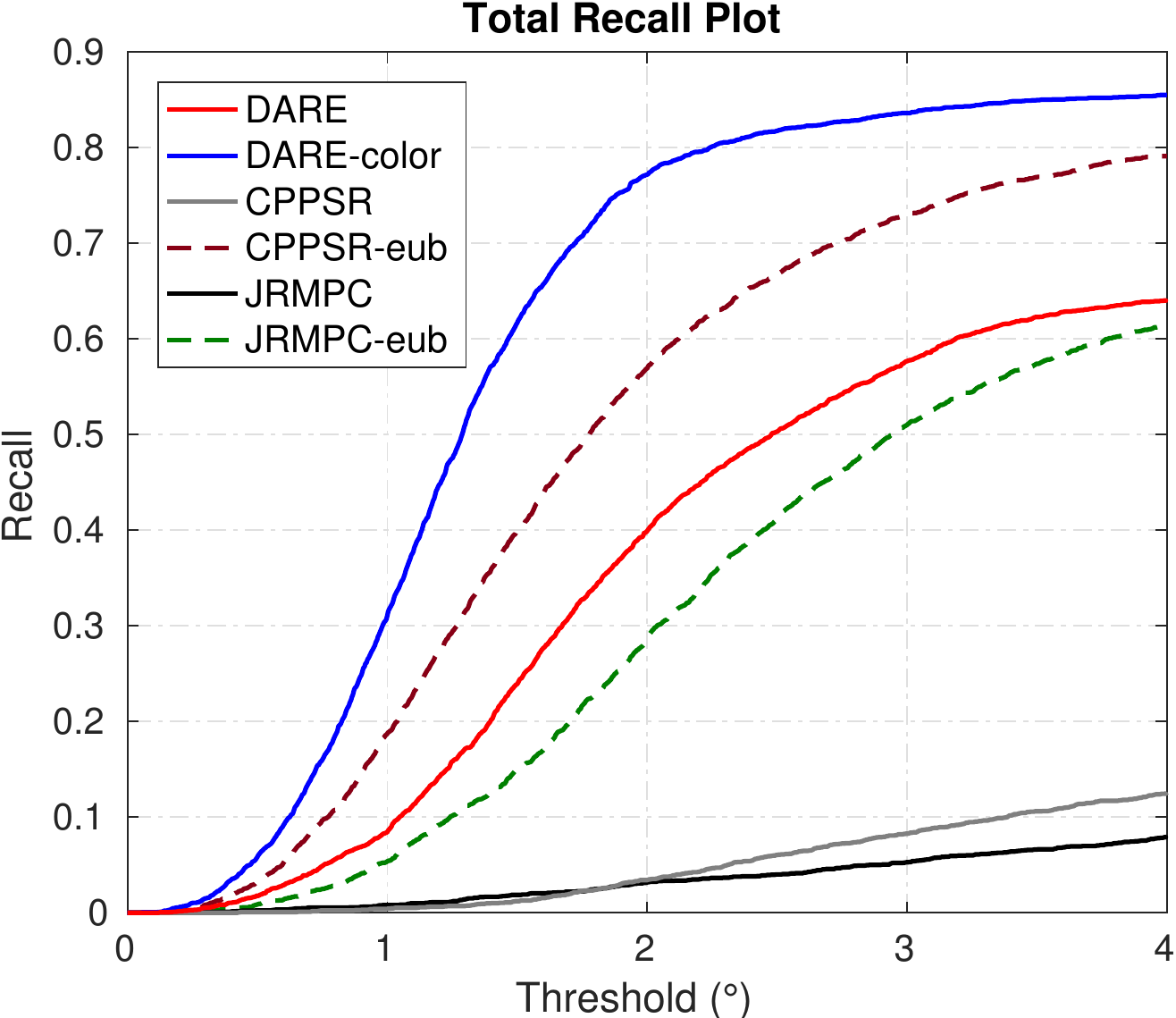}\vspace{-1mm}
	\caption{A multi-view registration comparison of our density adaptive model and existing methods, in terms of angular error on the VPS indoor dataset. Our model provides lower failure rate compared to the baseline methods JRMPC and CPPSR, also in comparison to the empirical upper bound.}\vspace{-2mm}
	\label{fig:joint_color}
\end{figure}

Table \ref{fig:joint_vpstls} shows pairwise registration comparisons in terms of angular error on the joint dataset. We compare the baseline JRMPC \cite{evangelidis2014generative} with both of our sampling density models: DARE and DARS. We also show the results for DARS without using normals, \ie setting $\gamma = 0$ in section~\ref{sec:sensor_model}, in the DARS-g0 curve. All the three variants of our density adaptive approach significantly improve over the baseline JRMPC \cite{evangelidis2014generative}. Further, our DARE model provides the best results. It significantly reduces the failure rate from 90.4$\%$ to 43.3$\%$, compared to the JRMPC method.

We also compare our empirical density adaptive model with several existing methods in the literature. Table~\ref{tab:full} shows the comparison of our approach with the JRMPC \cite{evangelidis2014generative}, ICP\footnote{We use the built-in Matlab implementation of ICP.} \cite{ICP_PAMI92}, and CPD \cite{myronenko2010point} methods. We present numerical values for the methods in terms of average inlier angular error and the failure rate. 

Additionally, we evaluate the existing methods using re-sampling. In the supplementary material we provide an evaluation of different re-sampling approaches at different sampling rates. For each of the methods JRMPC \cite{evangelidis2014generative}, ICP \cite{ICP_PAMI92}, and CPD \cite{myronenko2010point}, we select the best performing re-sampling approach and sampling rate. In practical applications however, such comprehensive exploration of the re-sampling parameters is not feasible. In this experiment, the selected re-sampling settings serve as empirical upper bounds, denoted by \emph{-eub} in the method names in table~\ref{tab:full}. 

From table~\ref{tab:full} we conclude that regardless of re-sampling approach, our DARE still outperforms JRMPC, both in terms of robustness and accuracy. The best performing method overall was the empirical upper bound for CPD with re-sampling. However, CPD is specifically designed for pairwise registration, while JRMPC and our approach also generalize to multi-view registration.

\subsection{Multi-view registration}
We evaluate our approach in a multi-view setting, by jointly registering all four point sets in the VPS indoor dataset. We follow a similar protocol as in the pairwise registration case (see supplementary material). In addition to the JRMPC, we also compare our color extension with the CPPSR approach of \cite{DanelljanCVPR2016}. Table~\ref{tab:multi} and figure \ref{fig:joint_color} shows the multi-view registration results on the VPS indoor dataset. As in the pairwise scenario, the selected re-sampled versions are denoted by \emph{-eub} in the method name. We use the same re-sampling settings for JRMPC and CPPSR as for JRMPC in the pairwise case. Both JRMPC and CPPSR have a significantly lower accuracy and a higher failure rate compared to our sampling density adaptive models. We further observe that re-sampling improves both JRMPC and CPPSR, however, not to the same extent as our density adaptive approach. Figure \ref{fig:multi} shows a qualitative comparison between our color based approach and the CPPSR method \cite{DanelljanCVPR2016}. In agreement with the pairwise scenario (see figure~\ref{fig:intro}) CPPSR locks on to the high density regions, while our density adaptive approach successfully registers all scans, producing an accurate reconstruction of the scene. Further, we provide additional results on the VPS outdoor dataset in the supplementary material.

\begin{table}
	\centering
	\resizebox{0.8\columnwidth}{!}{%
	\begin{tabular}{lcc}
		\toprule
		&Avg. inlier error ($^\circ$)&Failure rate (\%)\\\midrule
		CPPSR&2.57$\pm$0.837&87.4\\
		CPPSR\emph{-eub}&1.63$\pm$0.807&20.9\\
		JRMPC&2.38$\pm$1.01&92.1\\
		JRMPC\emph{-eub}&2.13$\pm$0.83&38.6\\\midrule
		{\bf DARE-color}&1.26$\pm$0.61&14.5\\
		{\bf DARE}&1.84$\pm$0.80&36.0\\\bottomrule
	\end{tabular}
	}\vspace{1mm}
	\caption{A multi-view registration comparison of our density adaptive model with existing methods in terms of average inlier angular error and failure rate on the VPS indoor dataset. Methods with \emph{-eub} in the name are empirical upper bounds. Our model provides improved results, both in terms of robustness and accuracy.}
	\label{tab:multi}
\end{table}

\section{Conclusions}
We investigate the problem of sampling density variations in probabilistic point set registration. Unlike previous works, we model both the underlying structure of the 3D scene and the acquisition process to obtain robustness to density variations. Further, we jointly infer the scene model and the transformation parameters by minimizing the KL divergence in an EM based framework. Experiments are performed on several challenging Lidar datasets. Our proposed approach successfully handles severe density variations commonly encountered in real-world applications.
\noindent\textbf{Acknowledgements}: This work was supported by the EU's Horizon 2020 Programme grant No 644839 (CENTAURO), CENIIT grant (18.14), and the VR grants: EMC2 (2014-6227), starting grant (2016-05543),  LCMM (2014-5928).

{\small
	\bibliographystyle{ieee}
	\bibliography{references}
}

\setcounter{equation}{0}
\setcounter{figure}{0}
\setcounter{table}{0}
\setcounter{section}{0}
\renewcommand{\theequation}{S\arabic{equation}}
\renewcommand{\thefigure}{S\arabic{figure}}
\renewcommand{\thetable}{S\arabic{table}}
\renewcommand{\thesection}{S\arabic{section}}
\onecolumn
\begin{center}
\textbf{\large Supplementary Material}
\end{center}
In this supplementary material we provide derivations of the proposed EM procedure and the observation weight function, based on both the empirical estimates and the sensor model. We also provide an evaluation of re-sampling methods and an analysis of the parameters introduced by our proposed model. Further, we present additional results and examples.
\section{Derivation of EM procedure}

We will here describe how the proposed objective in equation \eqref{eq:objective_q} in the paper, can be maximized using Expectation Maximization. To simplify the derivation, we first study the maximization of a single term $i$ in the objective \eqref{eq:objective_q} in the paper and drop the index $i$ to avoid clutter. We denote the likelihood in the local coordinate system of the point set as $p_{X}(x|\Theta) = p_{X}(x|\omega,\theta) = p_V(\phi(x;\omega)|\theta)$. The objective is then to maximize,
\begin{align}
\label{eq:objective_qs}
&\mathcal{E}(\Theta) = \int_{\reals^3} \!\log\left(p_V(\phi(x; \omega)|\theta)\right) \!\frac{q_V(\phi(x; \omega))}{q_{X}(x)} q_{X}(x) \,\diff x =  \int_{\reals^3} \!\log\left(p_X(x| \Theta)\right) \!\frac{q_V(\phi(x; \omega))}{q_{X}(x)} q_{X}(x) \,\diff x \,.
\end{align}
Here, $f(x) = \!\frac{q_V(\phi(x; \omega))}{q_{X}(x)}$ is the observation weight function. We now introduce a latent random variable $Z \in \{1, \ldots, K\}$ that assigns a 3D-point $V$ to a particular mixture component $k$. Using the relation $\log p_X(x|\Theta) = \log(p_{Z,X}(k,x|\Theta)) - \log(p_{Z|X}(k|x,\Theta))$, equation \eqref{eq:objective_qs} can be written as
\begin{align}
\label{eq:objective_q2}
&\mathcal{E}(\Theta) =\int_{\reals^3} \!\log\left(p_{Z,X}(k,x|\Theta)\right) \!f(x) q_{X}(x) \,\diff x +  \int_{\reals^3} \! -\log\left(p_{Z|X}(k|x,\Theta)\right) \! f(x) q_{X}(x) \,\diff x
\end{align}

Following the derivation in \cite{bishop2006pattern} we first take the expectation of both sides in equation \eqref{eq:objective_q2} with respect to a distribution $Z \sim \tilde{p}(k)$. We then add and subtract $\int_{\reals^3} \sum_k {\tilde p(k)} \log\left({\tilde p(k)}\right) f(x) q_{X}(x) \,\diff x$ on the right hand side of \eqref{eq:objective_q2}. Since the left hand side does not depend on $Z$, and ${\tilde p(k)}$ sums to one we get:

\begin{align}
\label{eq:objective_q3}
\mathcal{E}(\Theta) &= \int_{\reals^3} \underbrace{\sum^K_{k=1} {\tilde p(k)}\log\left(\frac{p_{Z,X}(k,x| \Theta)}{\tilde p(k)}\right)}_{=: \mathcal{L}({\tilde p},\Theta)} f(x) q_{X}(x) \,\diff x +  \int_{\reals^3} \! \underbrace{-\sum_{k=1}^{K} {\tilde p(k)}\log\left(\frac{p_{Z|X}(k|x, \Theta)}{\tilde p(k)}\right)}_{= \text{KL}({\tilde p}||p)} f(x) q_{X}(x) \,\diff x \nonumber \\
&= \int_{\reals^3} \! \mathcal{L}({\tilde p},\Theta) f(x) q_{X}(x) \,\diff x+ \int_{\reals^3}\text{KL}({\tilde p}||p) f(x) q_{X}(x) \,\diff x\,,
\end{align}
where $\text{KL}({\tilde p}||p)$ is the KL divergence from ${\tilde p}$ to the posterior distribution $p_{Z|X}(k|x, \Theta)$. We know that $\text{KL}({\tilde p}||p) \geq 0$ with equality if and only if ${\tilde p}(k)=p_{Z|X}(k|x, \Theta)$. Hence, $\int_{\reals^3} \! \mathcal{L}({\tilde p},\Theta) f(x) q_{X}(x) \,\diff x$ is a lower bound of $\mathcal{E}(\Theta)$. In the E-step we maximize the lower bound by setting ${\tilde p}=p$ given the current parameters $\Theta^n$. This leads to equality between the objective $\mathcal{E}(\Theta)$ and the lower bound at the current parameter estimate $\mathcal{E}(\Theta^n) = \mathcal{Q}(\Theta^n,\Theta^n) + D$. Here, $D = \int_{\reals^3} -\sum^K_{k=1} {\tilde p(k)}\log\left(\tilde p(k)\right) f(x) q_{X}(x) \,\diff x$, and with ${\tilde p}(k)=p_{Z|X}(k|x, \Theta^n)$, $\mathcal{Q}(\Theta,\Theta^n)$ is obtained as

\begin{align}
\label{eq:mstep}
\mathcal{Q}(\Theta,\Theta^n) &= \int_{\reals^3} \sum_{k=1}^{K} p_{Z|X}(k|x, \Theta^n) \log\left(p_{Z,X}(k,x| \Theta)\right)f(x) q_{X}(x) \,\diff x \nonumber \\
&= \int_{\reals^3} \!E_{Z|x,\Theta^n} \left[\log\left(p_{X,Z}(x,Z|\Theta)\right) \right] f(x) q_{X}(x) \,\diff x \,.
\end{align}
In the M-step we maximize the lower bound with respect to $\Theta$ to update the parameters by $\Theta^{n+1} = \argmax_\Theta \mathcal{Q}(\Theta,\Theta^n)$.

Since the exact value of the integral in \eqref{eq:mstep} is intractable, we treat the observations $x_j$ as a Monte Carlo sampling. This results in an approximation of the lower bound,
\begin{equation}
\label{eq:mstep2}
\mathcal{Q}(\Theta,\Theta^n) \approx Q(\Theta,\Theta^n) =
 \frac{1}{N} \sum_{j=1}^{N} \sum_{k=1}^{K} \alpha_{jk}^n f(x_{j}) \log\left(p_{X,Z}(x_{j},k|\Theta)\right) \,,
\end{equation}
where $\alpha_{jk}^n = p_{Z|X}(k|x_{j},\Theta^n)$. As described in the paper, with independent point samples we obtain the latent posteriors as
\begin{equation}
\label{eq:latent_posteriors}
p_{Z|X}(k|x,\Theta) = \frac{p_{X,Z}(x,k|\Theta)}{\sum_{k=1}^{K} p_{X,Z}(x,k|\Theta)} = \frac{\pi_k \norm (\phi(x;\omega_k); \mu_k, \Sigma_k)}{\sum_{l=1}^K \pi_l \norm (\phi(x;\omega_l); \mu_l, \Sigma_l)} \,.
\end{equation}
Maximizing the lower bound will cause $\mathcal{E}(\Theta)$ to increase unless it is at a maximum. Note that, during the whole procedure described above, we only evaluated the observation weight function $f(x)$ at the Monte Carlo sampling of $\mathcal{Q}(\Theta,\Theta^n)$. Although, $f(x)$ affects the maximization of $\mathcal{Q}(\Theta,\Theta^n)$, we see that it does not influence the derivation of the EM algorithm.

The derivation can trivially be generalized to multiple point sets.

In this case, the M-step extends to,
\begin{align}
\label{eq:EM2s}
\mathcal{Q}(\Theta,\Theta^n) \approx Q(\Theta,\Theta^n) =\sum_{i=1}^{M} \frac{1}{N_i} \sum_{j=1}^{N_i} \sum_{k=1}^{K} \alpha_{ijk}^n f_i(x_{ij}) \log\left(p_{V,Z}(\phi(x_{ij}; \omega_i),k|\theta)\right) \,.
\end{align}
We then apply the optimization procedure proposed in \cite{evangelidis2014generative} to maximize \eqref{eq:EM2s}.

\section{Derivation of sensor model}
Here, we derive the expression for the sensor model described in section \ref{sec:sensor_model} in the paper. We denote the measures $q_V(A) = \mathbb{P}(V \in A)$ and $q_{X_i}(A) = \mathbb{P}(V_i \in A)$ for the latent scene distribution $q_V$ and the sampling density $q_{X_i}$ respectively. By changing the variables in the integral to the reference frame $v = \phi_i(x)$, the measure $q_{X_i}(A)$ can be written as,
\begin{equation}
\label{eq:sensor_model1}
q_{X_i}(A) = \int_{S_i\cap A} \! \frac{g_i}{|S|} \, \diff S_i = \int_{\phi_i(S_i\cap A)} g_i \circ \phi^{-1}_i \, \frac{\diff S}{|S|} = \int_{S \cap \phi_i(A)} g_i \circ \phi^{-1}_i \, \frac{\diff S}{|S|} \,.
\end{equation}
Here, we have used the fact that $\phi_i(x) = R_i x + t_i$ is an isometric bijection.

From the definition $q_V(A) = \frac{1}{|S|}\int_{S \cap A} \diff S$, we see that $\frac{\diff q_V}{\diff S} = \frac{1}{|S|}$ on $S$ and zero elsewhere. From \eqref{eq:sensor_model1} we thus obtain,
\begin{equation}
\label{eq:sensor_model2}
q_{X_i}(A) = \int_{\phi_i(A)} g_i \circ \phi^{-1}_i \frac{\diff q_V}{\diff S} \diff S =  \int_{\phi_i(A)} g_i \circ \phi^{-1}_i \diff q_V = \int_A g_i \,\diff (q_V \circ \phi_i)\,.
\end{equation}
In the last equality in \eqref{eq:sensor_model2}, we have performed another change of variables back to the sensor-based coordinate frame. Here, $q_V \circ \phi_i(A) = q_V(\phi_i(A))$ denotes the composed measure.

Since, $q_V$ is derived from the Lebesgue measure on the surface $dS$, it is $\sigma$-finite \cite{durrett2010probability}. Furthermore, we also see that $q_V \circ \phi_i$ is absolutely continuous with respect to $q_{X_i}$, since the definition of $g_i$ (eq.~\eqref{eq:sensor_model} in the paper) guarantees that $q_{X_i}(A) = 0 \implies q_V \circ \phi_i(A) = 0$. We also see that $q_{X_i}(A)$ is $\sigma$-finite since $g_i$ is bounded everywhere except for the singularity in the origin of the sensor reference frame (the sensor position). The singularity of $g_i$ in the origin is not a problem if we assume that $S_i$ do not intersect the sensor center. From \eqref{eq:sensor_model2} we obtain the Radon-Nikodym derivative $\frac{\diff q_{X_i}}{\diff (q_V \circ \phi_i)} = g_i$. We can thus conclude, from the properties of the Radon-Nikodym derivative \cite{durrett2010probability}, that $f_i = \frac{\diff (q_V \circ \phi_i)}{\diff q_{X_i}} = \frac{1}{g_i}$.

\section{Derivation of empirical sample model}
Here, we add more details to the description of the empirical sensor model in section 3.4.2. As it is sufficient to study a single point set, we drop the index $i$.
The observation weight function that we estimate is $f(x) = \frac{q_V(\phi(x))}{q_X(x)}$. In the local neighborhood of a point on a surface $\bar{v} \in S$ we approximate the latent scene distribution as a Gaussian distribution in the normal direction
\begin{equation}
\label{eq:ng}
q_V(v) \approx \frac{1}{|S|} \norm ( \hat{n}_{\bar{v}}\tp(v - \bar{v}); 0, \sigma_{\hat{n}}^2(\bar{v})) \,.
\end{equation}
Assuming a rigid transformation $v = \phi(x) = R x + t$ we can write \eqref{eq:ng} as
\begin{align}
\label{eq:ng2}
 q_V(\phi(x)) &\approx \frac{1}{|S|} \norm ( \hat{n}_{\phi(\bar{x})}\tp(\phi(x) - \phi(\bar{x})); 0, \sigma_{\hat{n}}^2(\phi(\bar{x}))) = \frac{1}{|S|} \norm ( {\underbrace{(R\tp\hat{n}_{\phi(\bar{x})})}_{\hat{n}_x}}\tp(x - \bar{x}); 0, \sigma_{\hat{n}}^2(\phi(\bar{x}))) = \\
&=  \frac{1}{|S| \sqrt{2 \pi \sigma_{\hat{n}_{\bar{x}}}^2(\bar{x})}} e^{- \frac{1}{2\sigma_{\hat{n}_{\bar{x}}}^{2}(\bar{x})}(x - \bar{x})\tp \hat{n}_x \hat{n}_x\tp(x - \bar{x})}\,,
\end{align}
where $\hat{n}_{\bar{x}}$ is the surface normal vector at $\bar{x}$ in the reference frame of the point set $\mathcal{X}$.

For each 3D point $x$ in the point cloud we approximate the sampling density as a Gaussian distribution from the $L$ nearest neighbors
\begin{equation}
\label{eq:sg}
q_X(x) \approx \frac{L}{N} \norm ( x; \bar{x}, C) \,.
\end{equation}
Here, the covariance matrix is calculated as $C = \frac{1}{L-1} \sum_{j} (x_j-\bar{x})(x_j-\bar{x})\tp $ and $\bar{x}$ is the mean of the $L$ nearest neighbors. We now perform an eigenvalue decomposition of $C$ to obtain
\begin{align}
C = B D B\tp = (\hat{b}_1, \hat{b}_2, \hat{b}_3)
\begin{pmatrix}
\sigma_1^2&0&0 \\ 0&\sigma_2^2&0 \\ 0&0&\sigma_3^2
\end{pmatrix}
(\hat{b}_1, \hat{b}_2, \hat{b}_3)\tp = \sum_{i=1}^3 \sigma_i^2 \hat{b}_i \hat{b}_i\tp \quad \text{and} \\
C ^{-1} = (\hat{b}_1, \hat{b}_2, \hat{b}_3)
\begin{pmatrix}
\sigma_1^{-2}&0&0 \\ 0&\sigma_2^{-2}&0 \\ 0&0&\sigma_3^{-2}
\end{pmatrix}
(\hat{b}_1, \hat{b}_2, \hat{b}_3)\tp  = \sum_{i=1}^3 \sigma_i^{-2} \hat{b}_i \hat{b}_i\tp\,,
\end{align}
where the eigenvalues are sorted in descending order. The sample distribution in \eqref{eq:sg} can be expanded to
\begin{equation}
\label{eq:qxn}
q_X(x) = \frac{L}{N \sqrt{2 \pi \det C}} e^{-\frac{1}{2}(x - \bar{x})\tp C^{-1} (x - \bar{x})} =  \frac{L}{N \sqrt{2 \pi \sigma_1^2 \sigma_2^2 \sigma_3^2}} e^{-\frac{1}{2}(x - \bar{x})\tp \left(\sum_{i=1}^3 \sigma_i^{-2} \hat{b}_i \hat{b}_i\tp\right)(x - \bar{x})}\,.
\end{equation}

The eigenvector $\hat{b}_3$ corresponding to the smallest eigenvalue
$\sigma_3$ approximates the surface normal, and the squared eigenvalue corresponds to the variance in this direction. We set $\hat{n}_{\bar{x}} = \hat{b}_3$ and $\sigma_{\hat{n}}^2(\bar{v}) = \sigma_3^2$. By inserting this into \eqref{eq:ng2} we get
\begin{align}
\label{eq:ng3}
& q_V(\phi(x)) \approx \frac{1}{|S| \sqrt{2 \pi \sigma_3^2}} e^{-\frac{1}{2}(x - \bar{x})\tp {\sigma_3^{-2}}\hat{b}_3 \hat{b}_3\tp(x - \bar{x})} \,,
\end{align}
The observation weight function can now be calculated from \eqref{eq:qxn} and \eqref{eq:ng3}
\begin{align}
f(x) &= \frac{q_V(\phi(x))}{q_X(x)} \approx \frac{N \sqrt{2 \pi \sigma_1^2 \sigma_2^2 \sigma_3^2}}{L |S| \sqrt{2 \pi \sigma_3^2}} e^{\frac{1}{2}(x - \bar{x})\tp \left(\sum_{i=1}^3 \sigma_i^{-2} \hat{b}_i \hat{b}_i\tp\right) (x - \bar{x})-\frac{1}{2}(x - \bar{x})\tp {\sigma_3^{-2}}\hat{b}_3 \hat{b}_3\tp(x - \bar{x})} = \nonumber \\
&=\frac{N \sigma_1 \sigma_2}{L |S|} e^{\frac{1}{2}(x - \bar{x})\tp \left(\sum_{i=1}^2 \sigma_i^{-2} \hat{b}_i \hat{b}_i\tp\right) (x - \bar{x})} = \frac{N \sigma_1 \sigma_2}{L |S|} e^{\frac{1}{2}(x - \bar{x})\tp \left(\sum_{i=1}^2 \sigma_i^{-2} \hat{b}_i \hat{b}_i\tp\right) (x - \bar{x})} \nonumber \\ &\propto \sigma_1 \sigma_2 e^{\frac{1}{2}(x - \bar{x})\tp \left(\sum_{i=1}^2 \sigma_i^{-2} \hat{b}_i \hat{b}_i\tp\right) (x - \bar{x})}\,.
\end{align}
This is equivalent to equation \eqref{eq:empirical_weights} in the paper.

\section{Experiments}
In this section we present more detailed results and analysis, complementary to the experiments provided in the main paper. We provide an evaluation of different re-sampling methods at varying sampling rate. Further, we analyze the impact of the parameters introduced by our proposed model. We present additional results for both pairwise and multi-view registration. 

For quantitative comparison between different methods we provide recall curves for both pairwise and multi-view registration. In addition to rotation error recall curves (see section~\ref{sec:exp_det} in the main paper for description), we also provide recall curves for the translation error. The translation error is calculated as the Euclidean norm of the difference between the ground truth and the found translation.

\subsection{Re-sampling evaluation}
\label{sec:rs}
\begin{figure}
	\centering
	\subfloat[Failure rate\label{fig:fps_failr}]{\includegraphics[width=0.4\columnwidth]{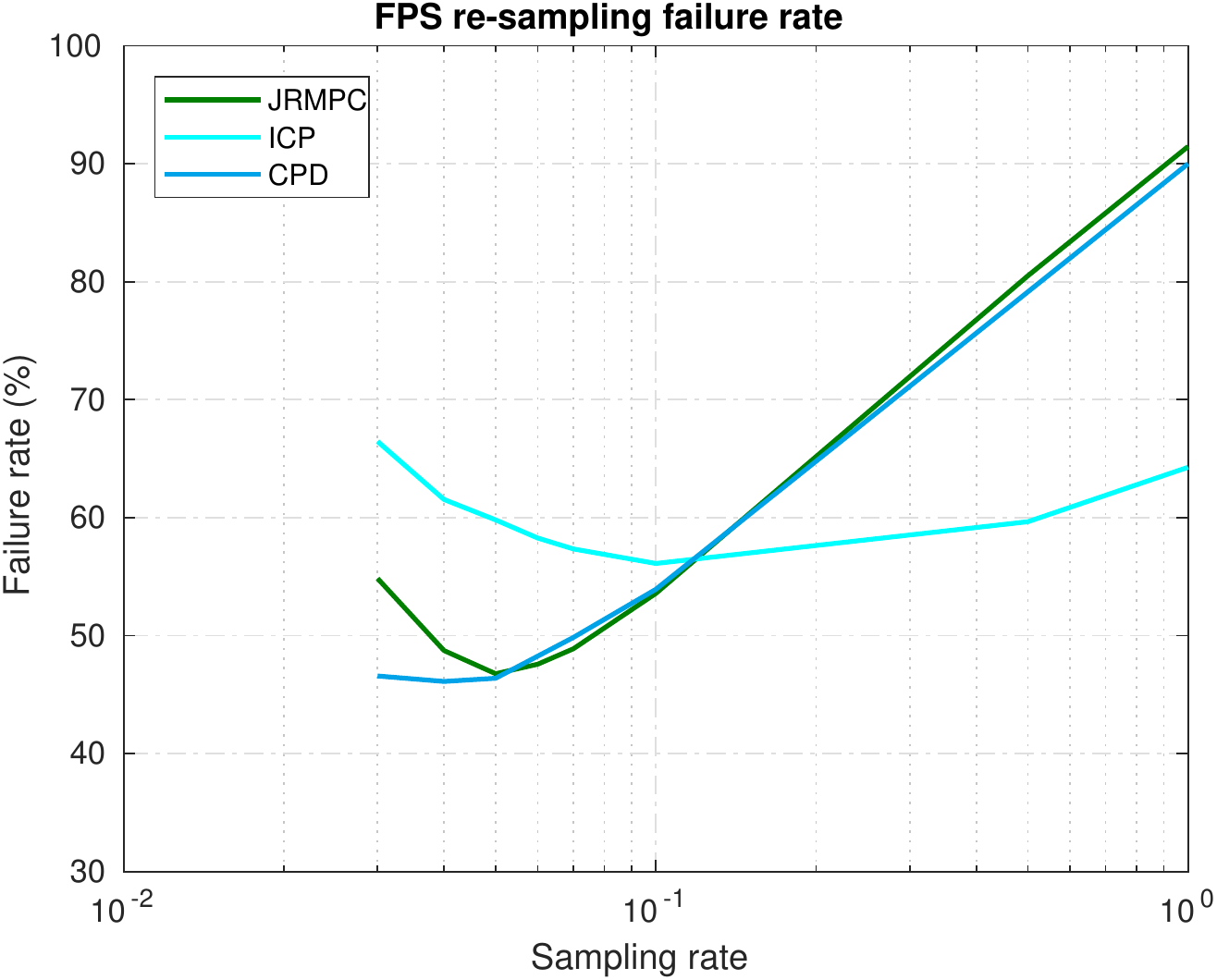}}\hspace{2mm}
	\subfloat[Inlier error\label{fig:fps_acc}]{\includegraphics[width=0.4\columnwidth]{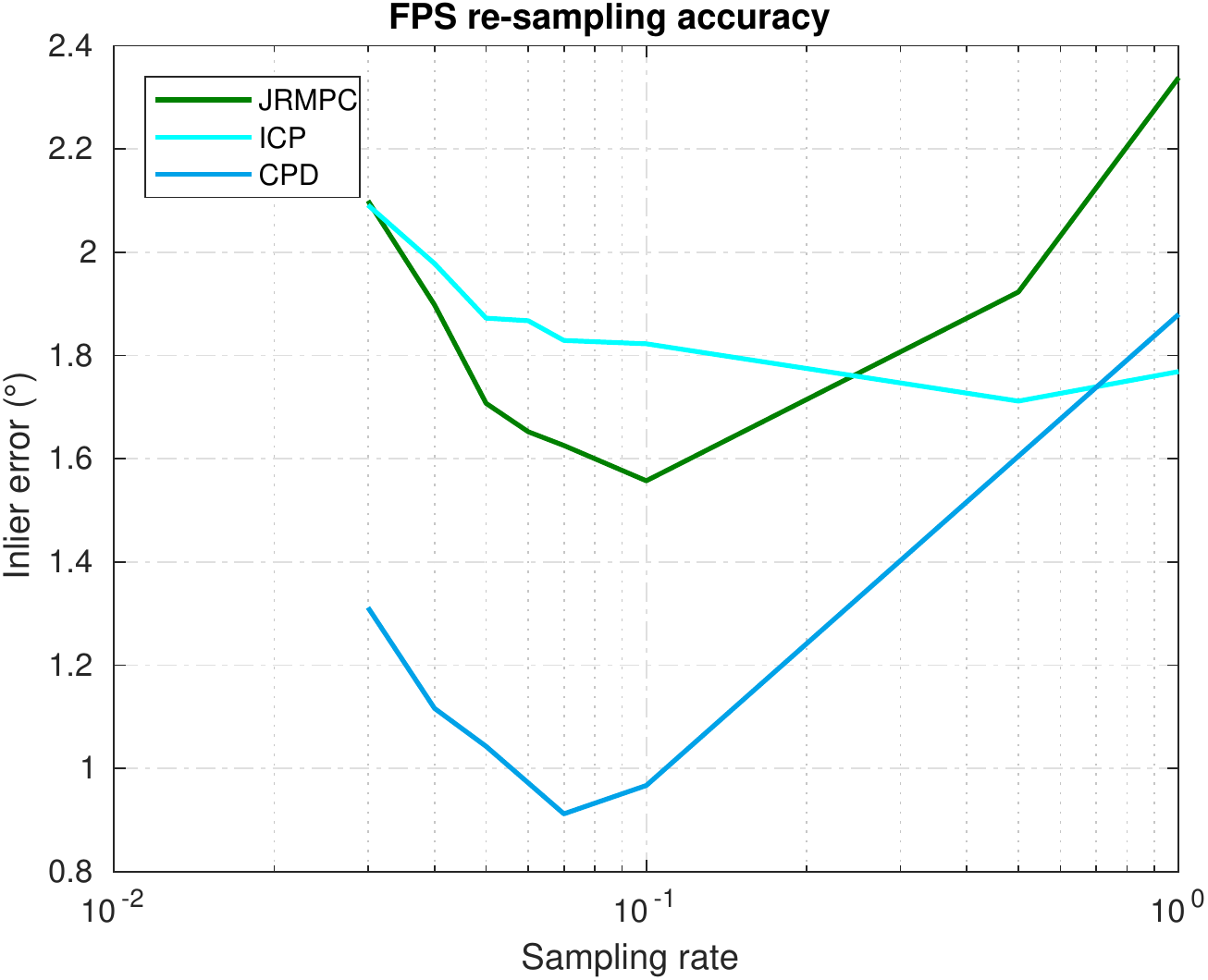}}
	\caption{Failure rates (a) and average inlier error (b) as a function of the sample rate using the FPS re-sampling method. }\vspace{-2mm}
	\label{fig:ds_fps}
\end{figure}

\begin{figure}
	\centering
	\subfloat[Failure rate\label{fig:grid_failr}]{\includegraphics[width=0.4\columnwidth]{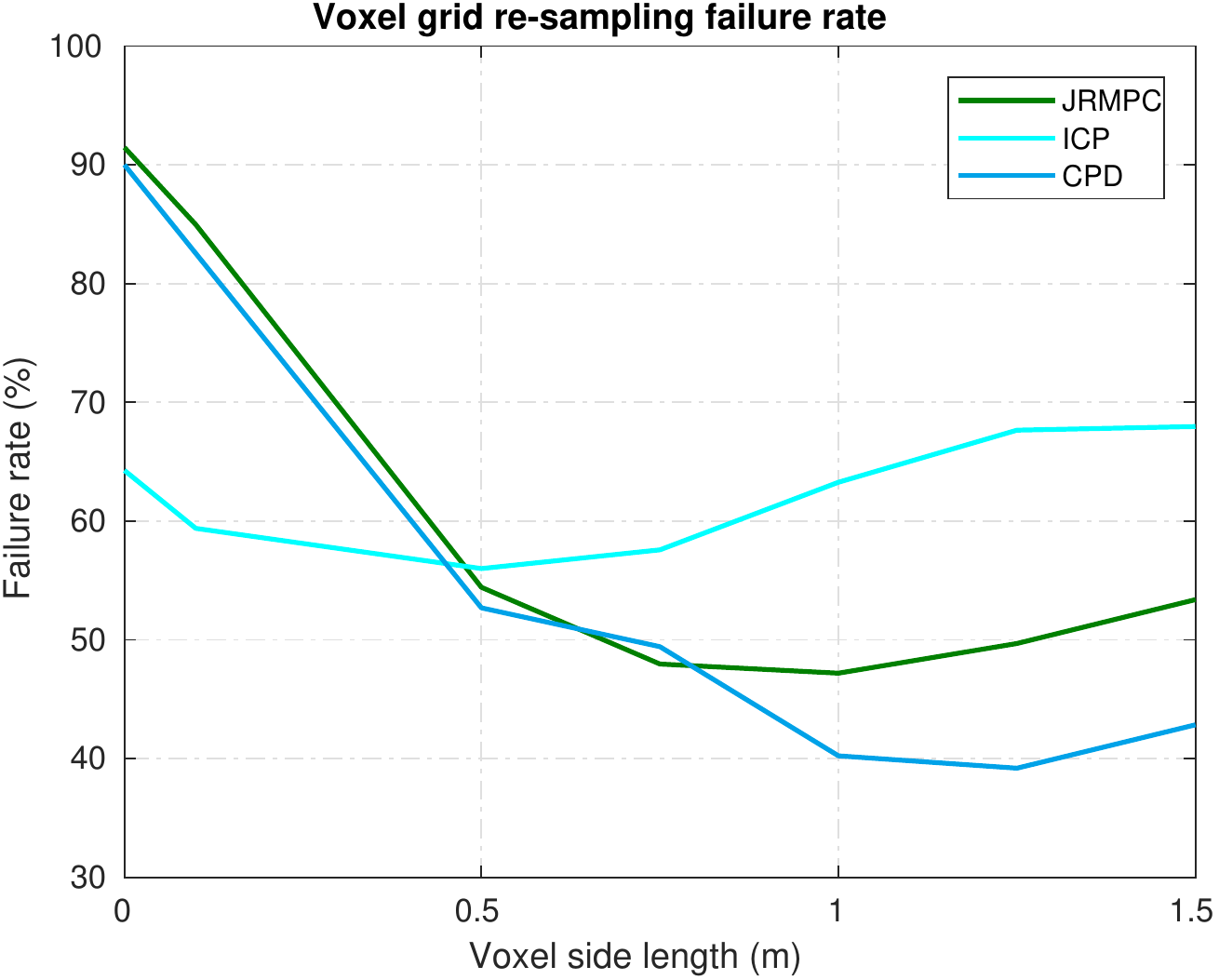}}\hspace{2mm}
	\subfloat[Inlier error\label{fig:grid_acc}]{\includegraphics[width=0.4\columnwidth]{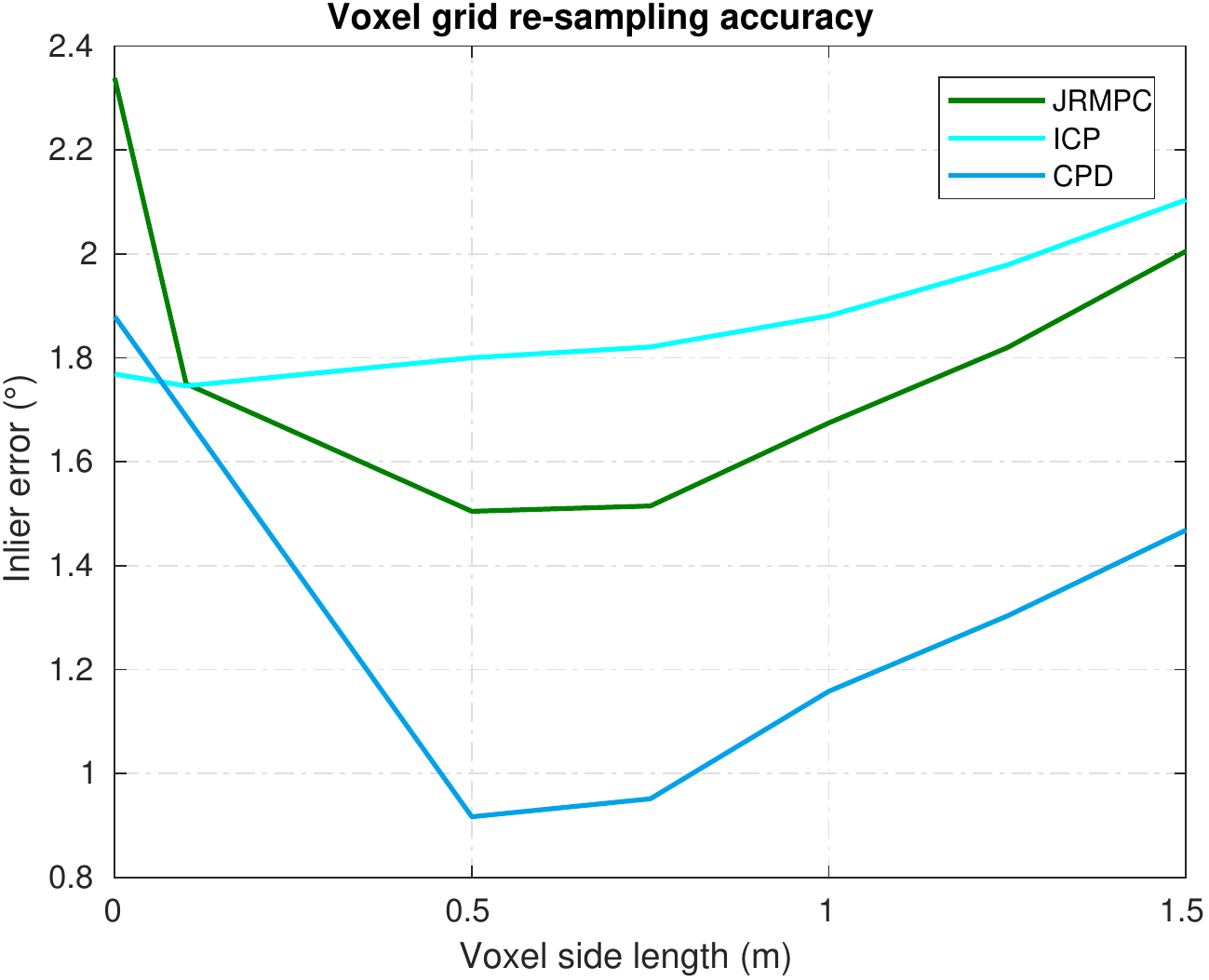}}
	\caption{Failure rates (a) and average inlier error (b) as a function of grid side length using the voxel grid re-sampling method.}\vspace{-2mm}
	\label{fig:ds_grid}
\end{figure}

\begin{figure}
	\centering
	\subfloat[Falure rate\label{fig:gss_failr}]{\includegraphics[width=0.4\columnwidth]{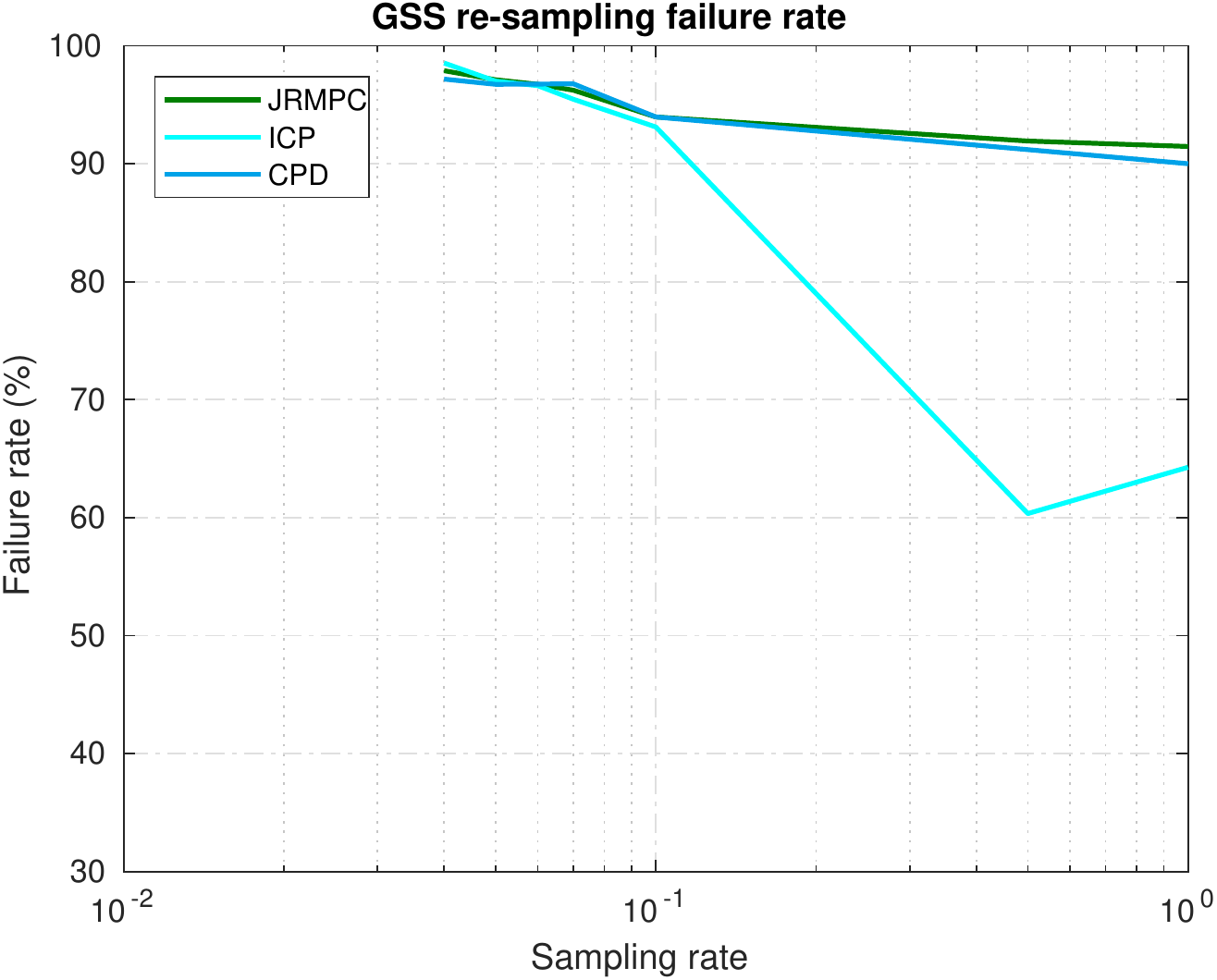}}\hspace{2mm}
	\subfloat[Inlier error\label{fig:gss_acc}]{\includegraphics[width=0.4\columnwidth]{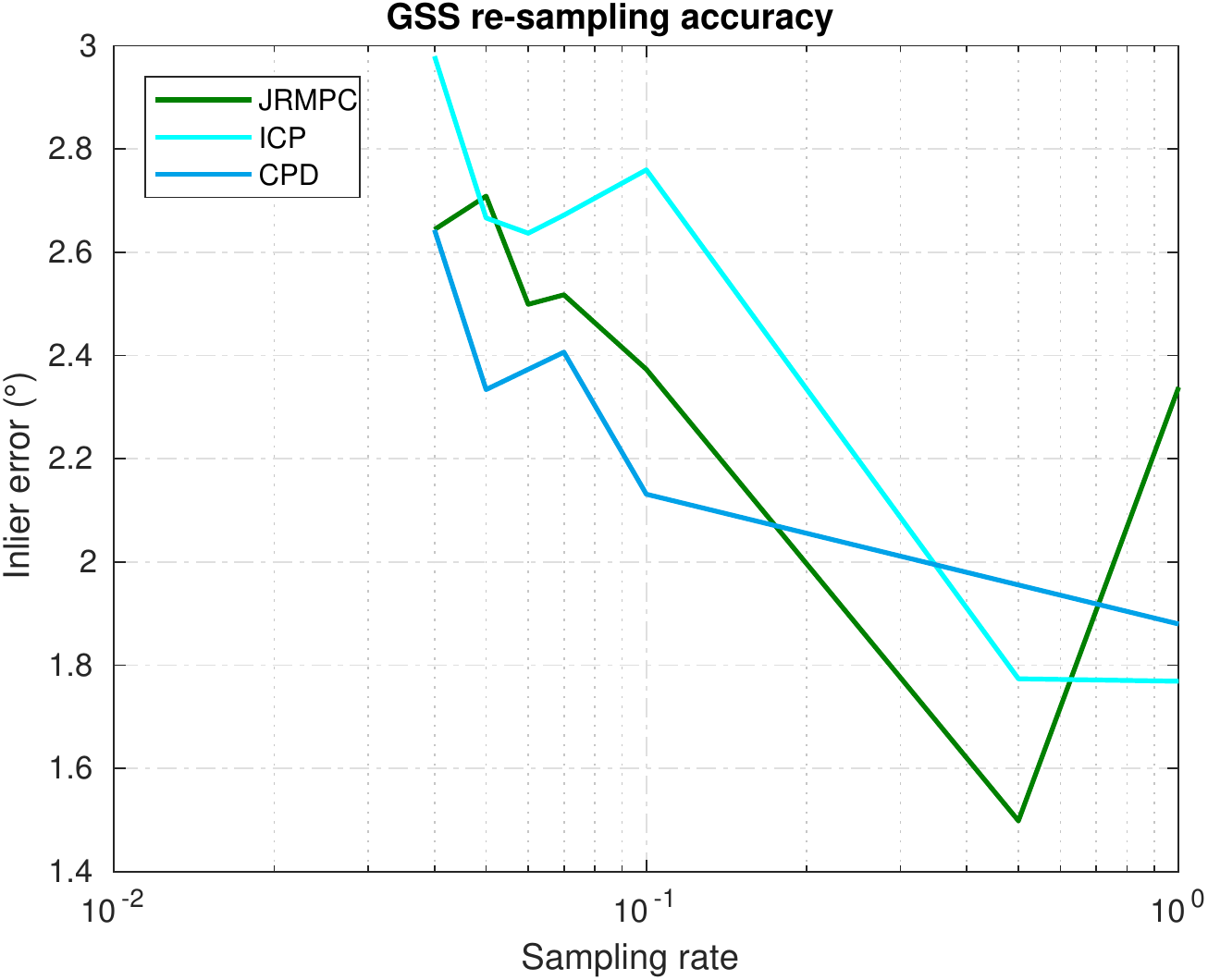}}
	\caption{Failure rates (a) and average inlier error (b) as a function of the sample rate using the GSS re-sampling method.}\vspace{-2mm}
	\label{fig:ds_gss}
\end{figure}

In the main paper we compare our approach to existing methods with re-sampling preprocessing. However, the selection of re-sampling technique is cumbersome and depends both on the dataset and the registration method. We evaluate JRMPC \cite{evangelidis2014generative}, ICP \cite{ICP_PAMI92} and CPD \cite{myronenko2010point}, using FPS \cite{eldar1997farthest}, GSS \cite{gelfand2003geometrically} and the voxel grid re-sampling at different sampling rates.  The evaluation is performed for pairwise registration on the same dataset as in the main paper, with a reduced number of pairs compared to the experiment in the main paper. This includes the facade and office datasets from TLS ETH and the indoor and outdoor dataset from VPS. In figure~\ref{fig:ds_fps}-\ref{fig:ds_gss}, we present the performance both in terms of failure rate and mean inlier error as a function over the sampling rate.

Figure \ref{fig:ds_fps} shows the performance of the registration methods using FPS re-sampling for different sampling rates. For JRMPC the robustness increases as the sampling rate decreases, with the lowest failure rate at sampling rate $0.05$ (see figure \ref{fig:fps_failr}). In \ref{fig:fps_acc} we see that the average inlier error increases as the sampling rate is significantly reduced. The best performance gains are observed for CPD, both in terms of robustness and accuracy, while ICP is only marginally affected.

Further, figure \ref{fig:ds_grid} shows the robustness and accuracy for the registration methods using voxel grid re-sampling. We varied the voxel side length between 0 to 1.5 meters. For JRMPC we observe similar performance as in the FPS case, with the lowest failure rate at a voxel side length of $1.0$ meters. As in the FPS case, the accuracy is degrading for low sampling rate (e.g large voxel side length). CPD strongly benefited from the voxel grid re-sampling for large voxel sizes, with the best performance recorded at 1.25 meters.

Finally, the GSS re-sampling method did not improve the robustness for JRMPC and CPD as we see in figure \ref{fig:ds_gss}. With a sampling rate at 0.5, we observe a small improvement in robustness for ICP.

From the results of the re-sampling evaluation we deduce that both the choice of sampling rate and the re-sampling approach have a significant impact on the performance of point set registration methods. The best performing re-sampling setting for JRMPC is FPS with a sampling rate of 0.05. For CPD and ICP the best performance is achieved using voxel grid re-sampling, with voxel side length 1.25 and 0.5 meters respectively. In the following experiments, we denote the methods using these re-sampling settings as the empirical upper bounds. 

\subsection{Parameter analysis}
Our density adaptive approach introduces two additional parameters to the JRMPC framework. First, we introduce a threshold $T$, which is used for clipping all observation weights larger than $T$ times the mean of the weights. This way we reduce the impact of potential outliers. Second, we introduce $L$, which is the number of neighbors used for calculating the empirical observation weights.

Table \ref{tab:param} shows the performance of our method in terms of average inlier error and failure rate for different values on $T$ and $L$. The experiment was performed for pairwise registration on the facade and office datasets from TLS ETH, the VPS indoor dataset, and the VPS outdoor dataset, with a reduced number of pairs compared to the experiment in the main paper. We see that by increasing $T$ the robustness is slightly improved, to the cost of lower accuracy. By increasing the number of neighbors $L$ the accuracy is improved, to the cost of reduced robustness. We set $T=8$ and $L=10$ for the experiments in the main paper.

In the experiments we have observed that our DARE approach occasionally benefits from re-sampling when $T$ is set to a lower value. This is expected, since a lower $T$-value leads to a behavior more similar to JRMPC.

\begin{table}
	\centering%
	\resizebox{0.3\columnwidth}{!}{%
		\begin{tabular}{lc|cc}
			\toprule
			$T$ &$L$&avg inlier err&failure rate \\\midrule
			6&10&1.49& 43.8\\
			8&10&1.49&42.3\\
			10&10&1.56&41.3\\
			12&10&1.59&42.0\\
			8&5&1.53&42.0\\
			8&20&1.46&42.8\\
			8&40&1.45&43.5\\\bottomrule
		\end{tabular}
	}\vspace{1mm}
	\caption{Analysis of the parameters $T$ and $L$ for our DARE method.  By increasing $T$ the robustness is slightly improved, to the cost of lower accuracy. By increasing the number of neighbors $L$ the accuracy can be improved, at the cost of reduced robustness. Overall, our method is not sensitive to the values of these parameters.}
	\label{tab:param}
\end{table}

\subsection{Pairwise Registration}
\label{sec:pairwise_s}
\begin{figure}
	\centering
	\subfloat[Rotation recall\label{fig:full}]{\includegraphics[width=0.4\columnwidth]{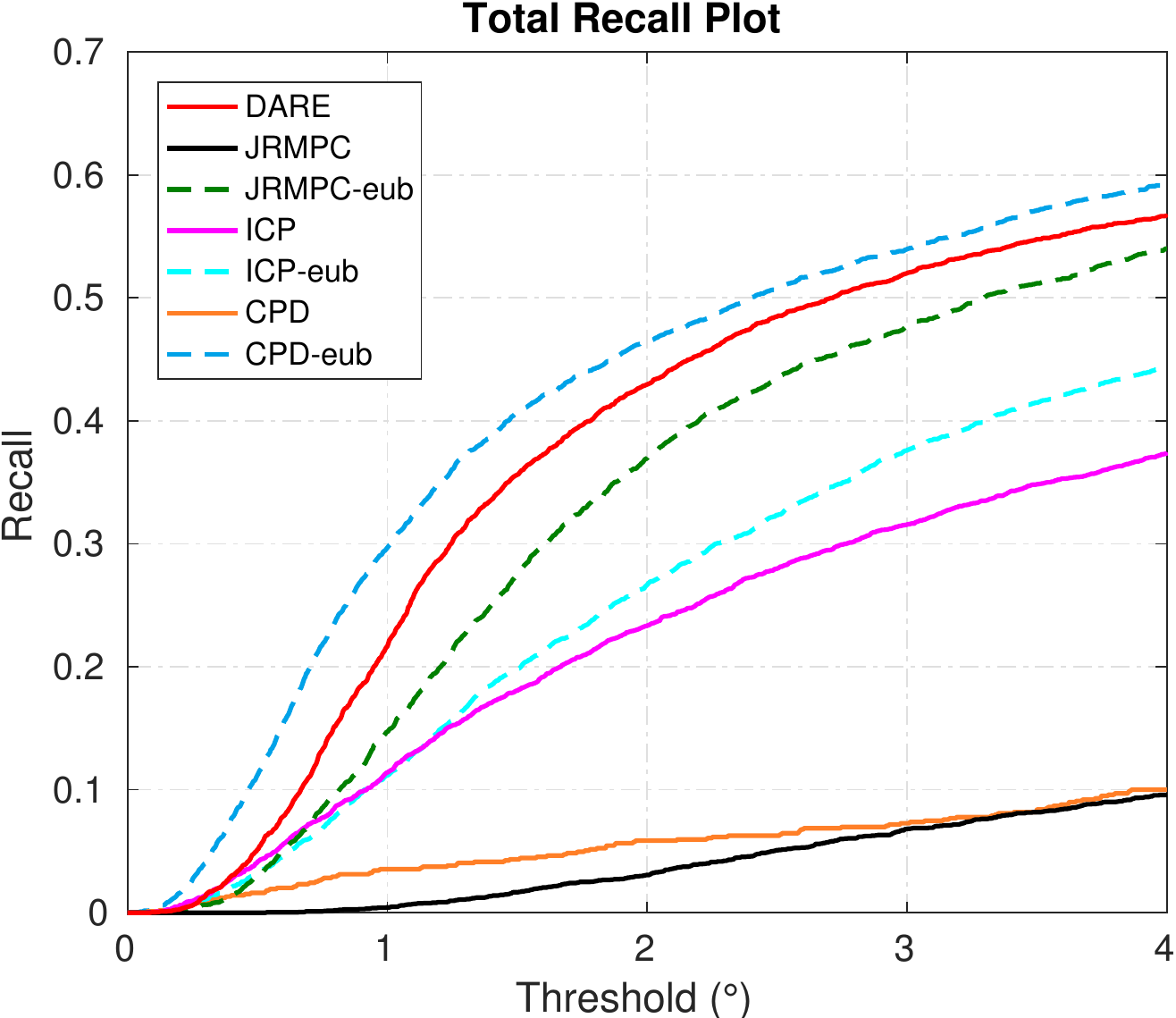}}\hspace{2mm}
	\subfloat[Translation recall\label{fig:full_t}]{\includegraphics[width=0.4\columnwidth]{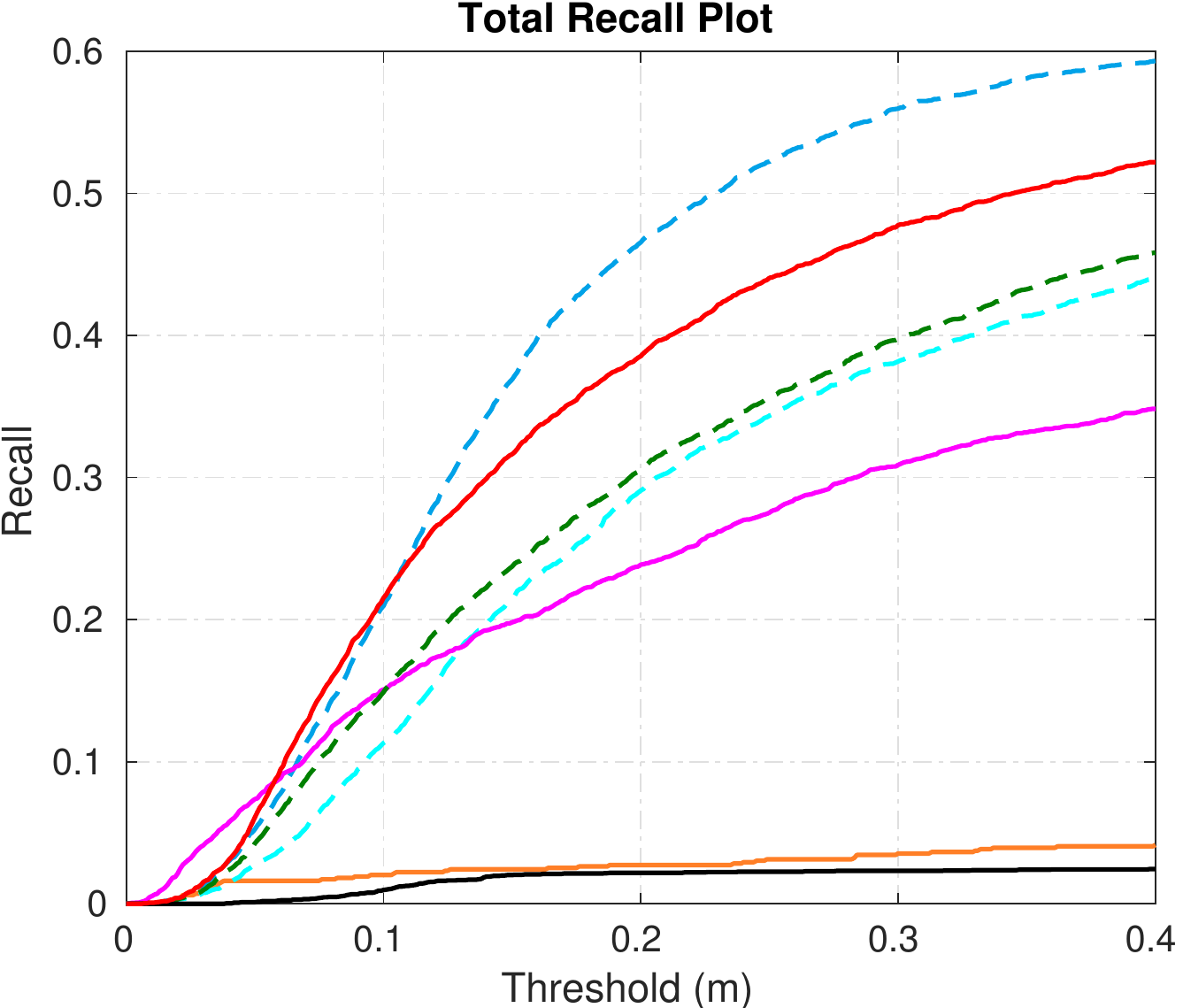}}
	\caption{Recall curves with respect to (a) the angular error and (b) translation error on the joint facade and office dataset in TLS ETH and the indoor and outdoor dataset in VPS. Our DARE method shows improved results with respect to the baseline JRMPC, with and without re-sampling  both in terms of rotation and the translation error.}\vspace{-2mm}
	\label{fig:full_cr}
\end{figure}

In table~\ref{tab:full} in the main paper we present numerical values for the methods in terms of average inlier angular error and the failure rate on the combined VPS and TLS ETH dataset. In figure \ref{fig:full_cr} we provide the corresponding recall plot. For all methods, the results are also presented using empirically optimized re-sampling for each method from the experiment in section \ref{sec:rs}. Methods using empirically optimized re-sampling denoted by adding \emph{-eub} in the method name. We see that our density adaptive method consistently improves over the baseline JRMPC with and without re-sampling, both in with respect to the rotation and translation error.

\begin{figure}
	\centering
	\subfloat[{\bf Facade}: Rotation recall \label{fig:fac}]{\includegraphics[width=0.24\columnwidth]{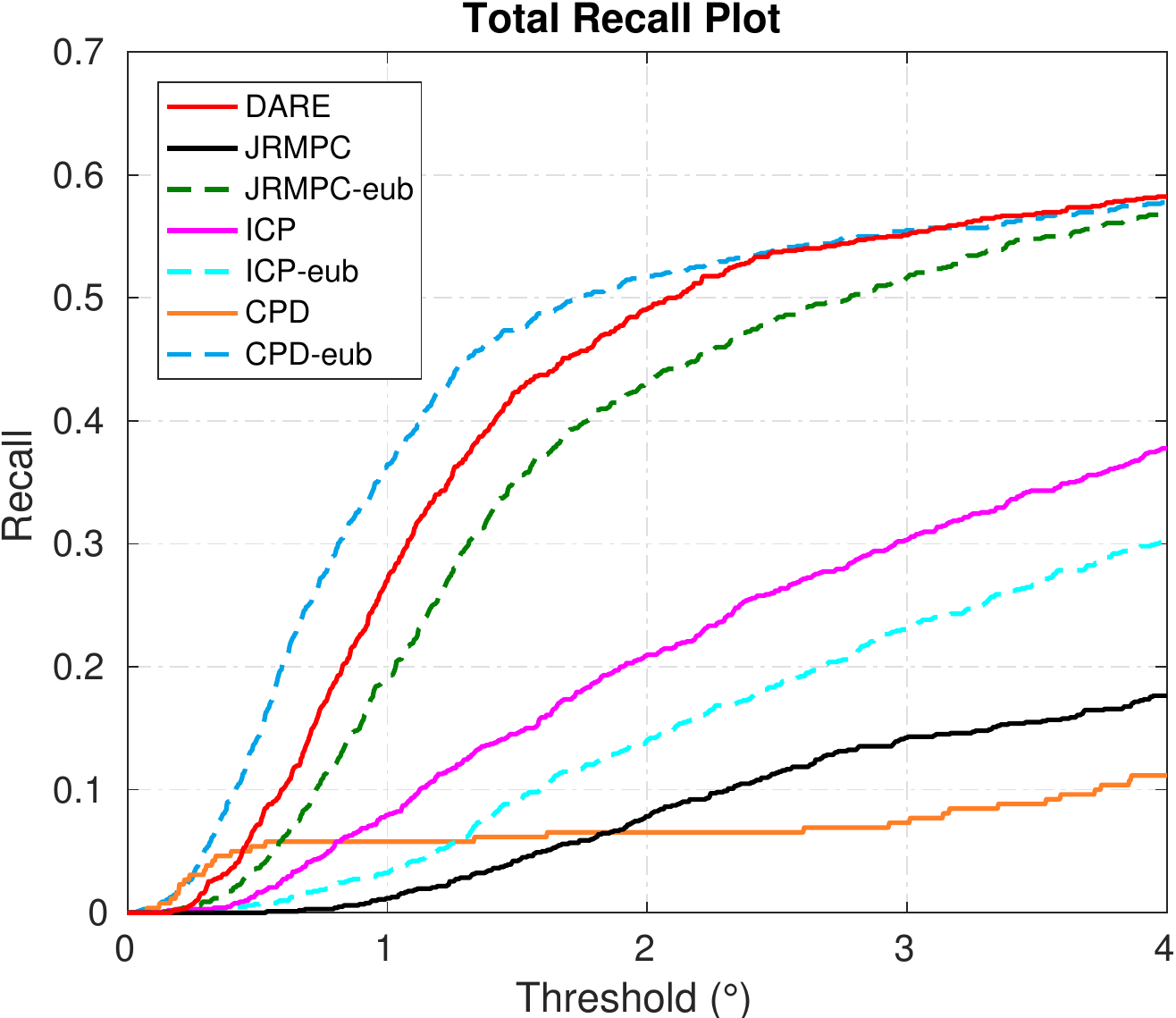}}
	\subfloat[{\bf Office}: Rotation recall \label{fig:off}]{\includegraphics[width=0.24\columnwidth]{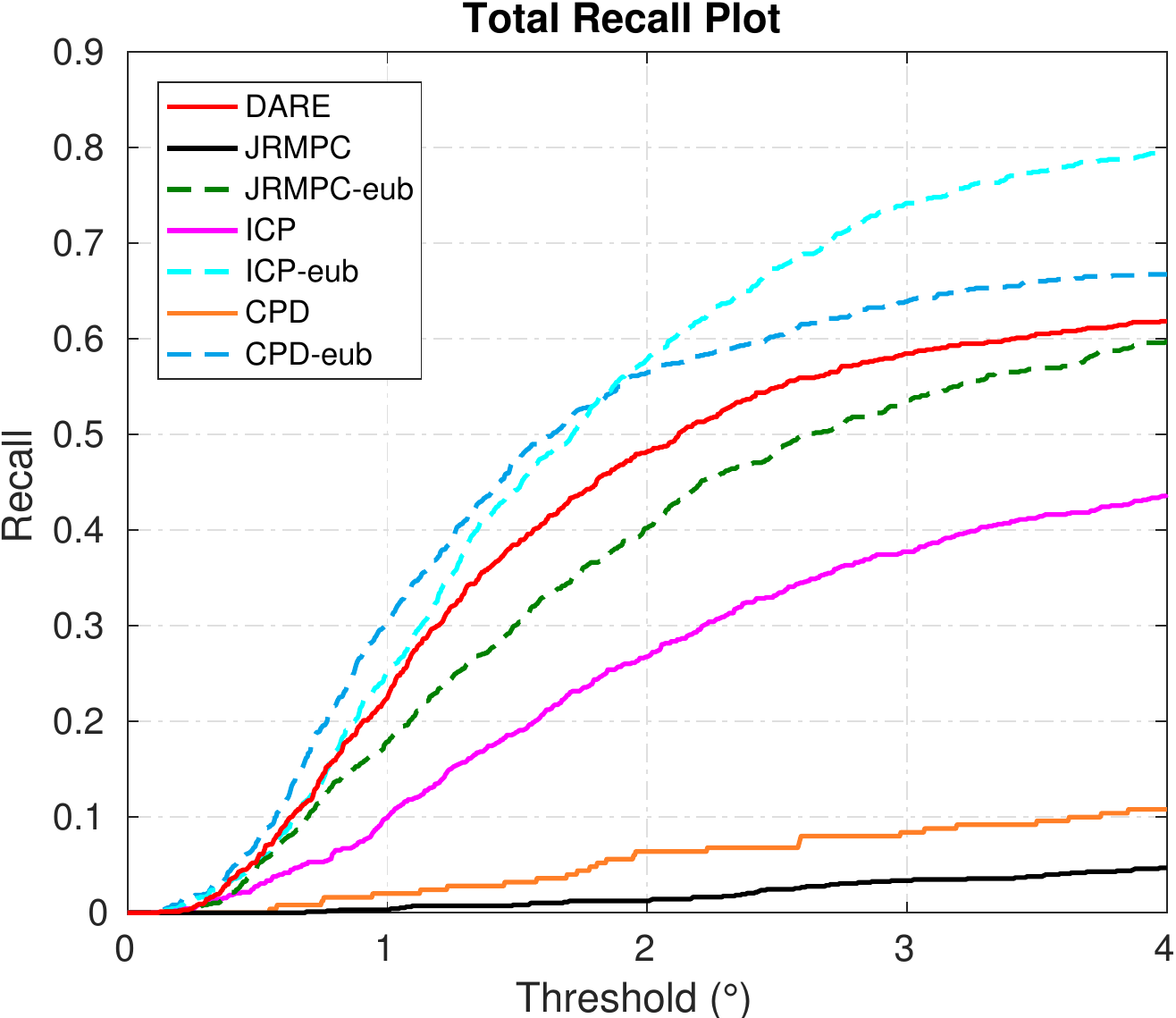}}
	\subfloat[{\bf VPS indoor}: Rotation recall\label{fig:vpsin}]{\includegraphics[width=0.24\columnwidth]{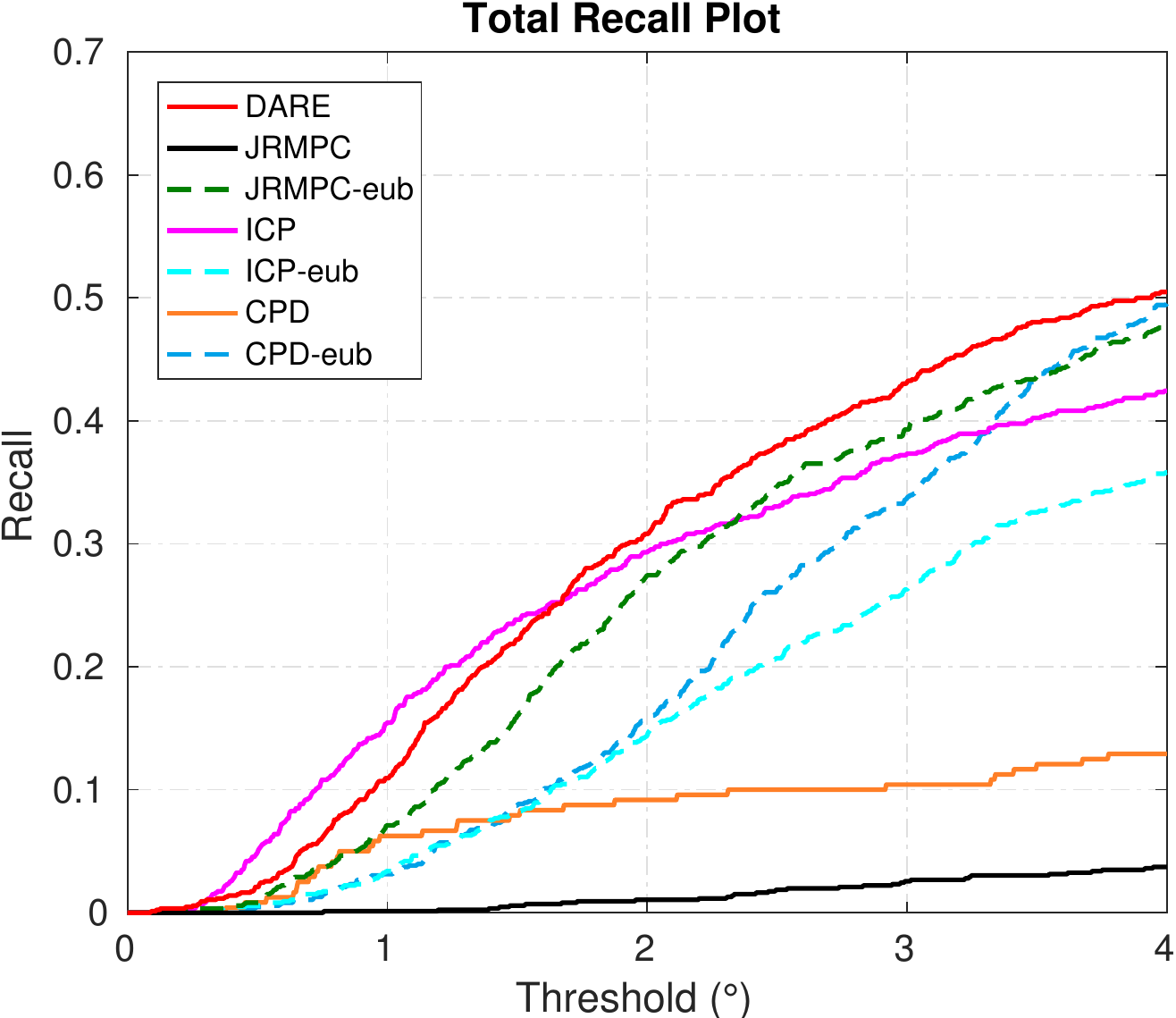}}
		\subfloat[{\bf VPS outdoor}: Rotation recall\label{fig:vpsout}]{\includegraphics[width=0.24\columnwidth]{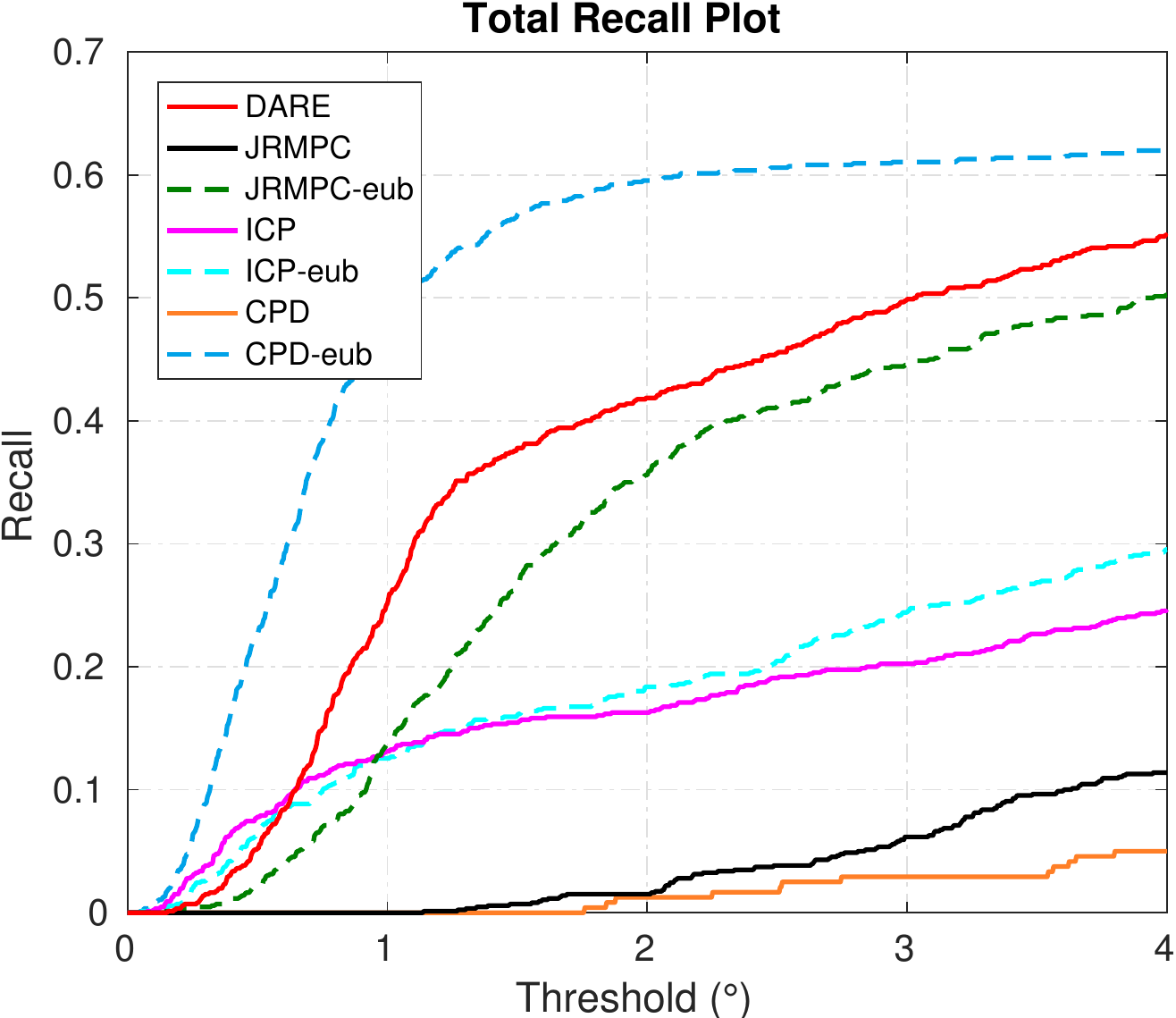}}\vspace{-2mm}\\
	\subfloat[{\bf Facade}: Translation recall\label{fig:fac_t}]{\includegraphics[width=0.24\columnwidth]{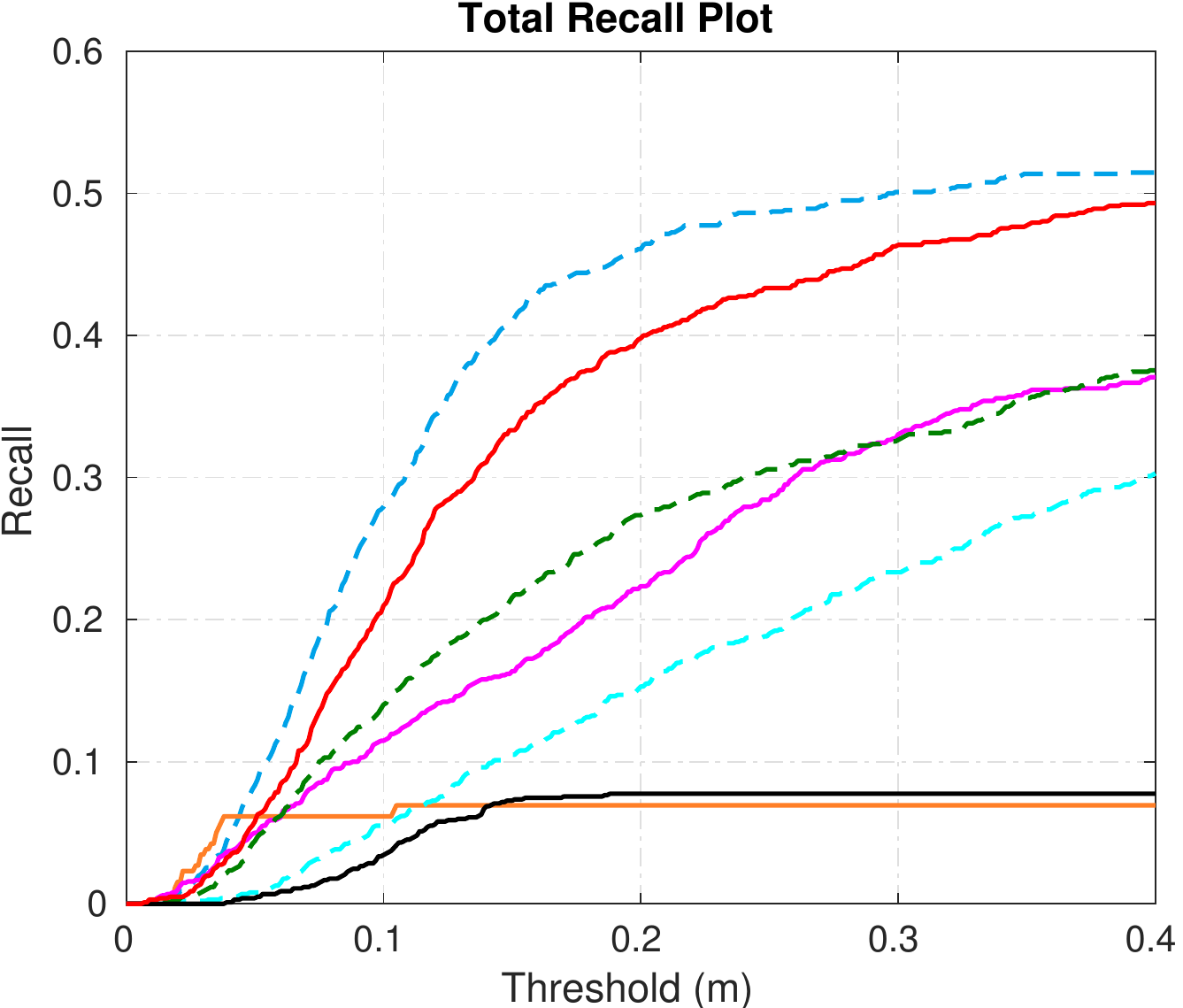}}
	\subfloat[{\bf Office}: Translation recall\label{fig:off_t}]{\includegraphics[width=0.24\columnwidth]{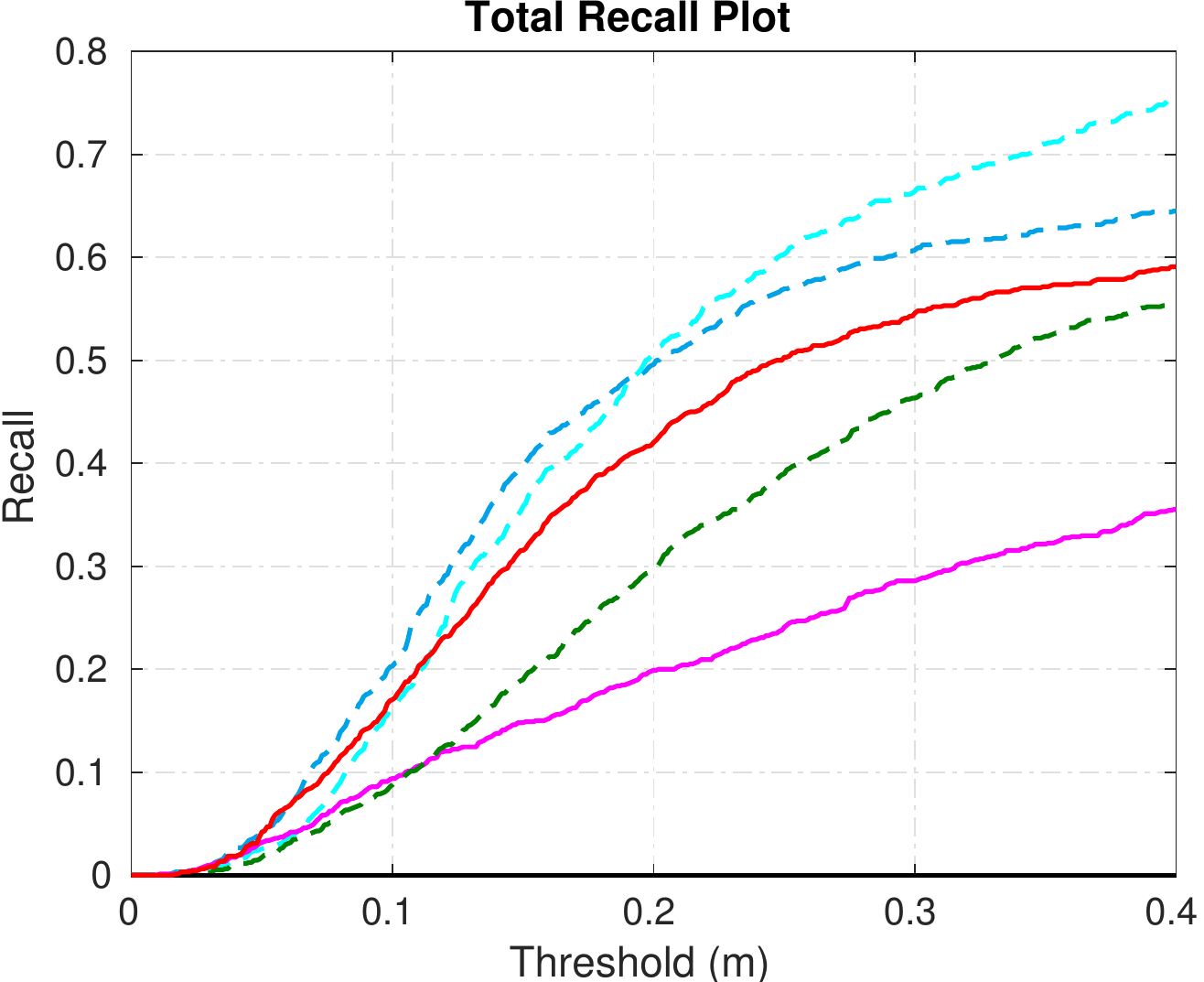}}
	\subfloat[{\bf VPS indoor}: Translation recall\label{fig:vpsin_t}]{\includegraphics[width=0.24\columnwidth]{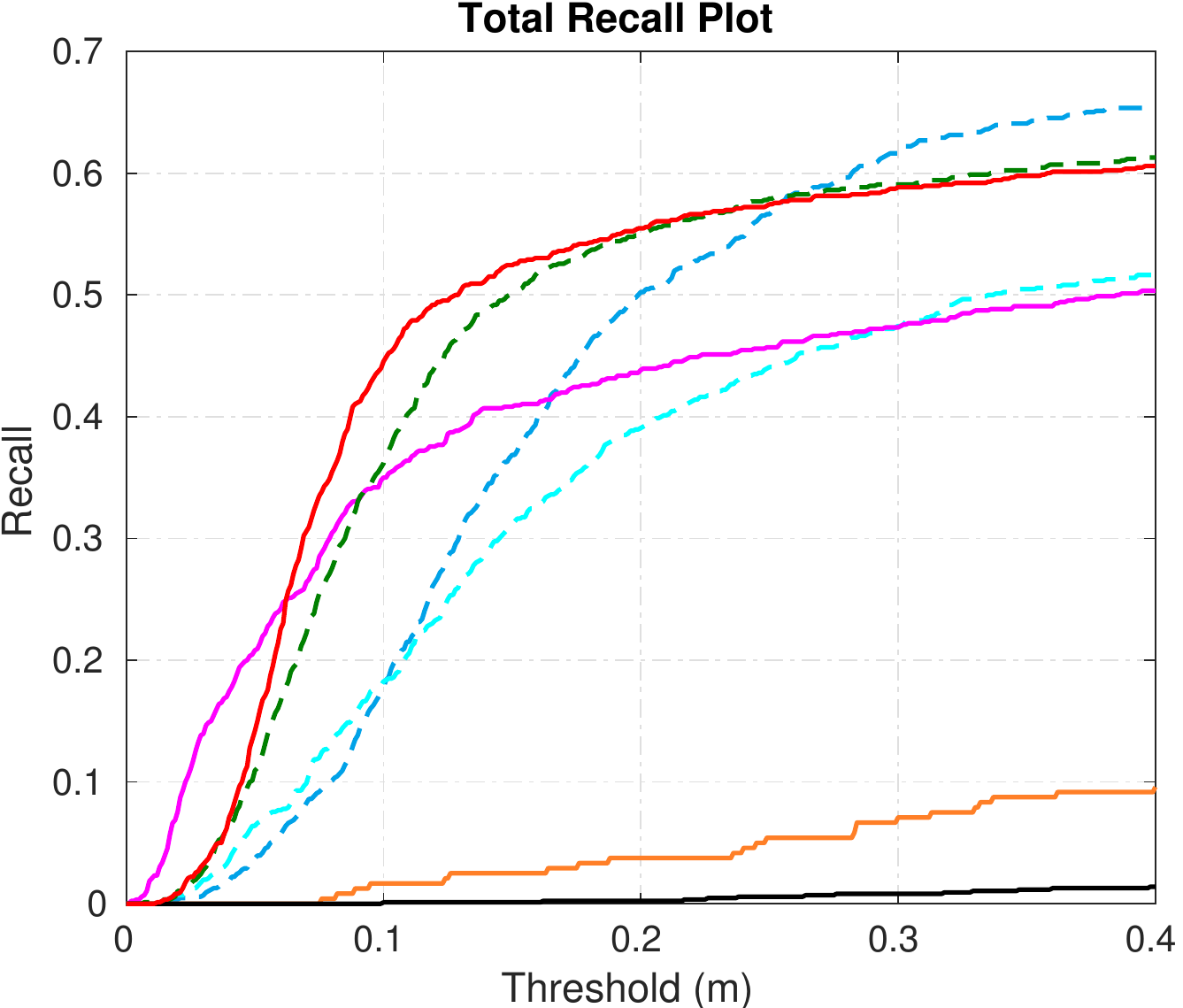}}
	\subfloat[{\bf VPS outdoor}: Translation recall\label{fig:vpsout_t}]{\includegraphics[width=0.24\columnwidth]{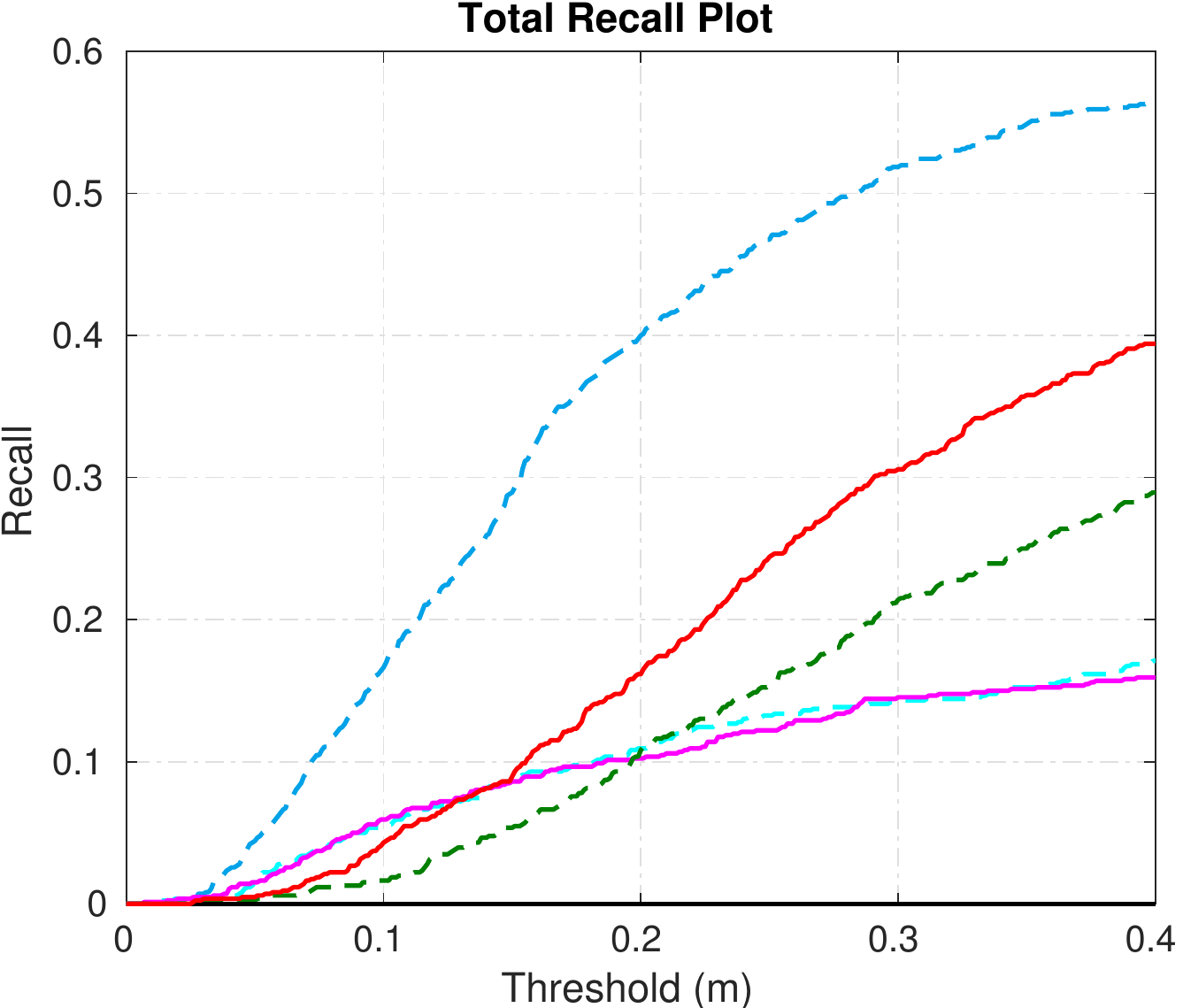}}
	\caption{Recall curves for the individual dataset with respect to the rotation error (top row) and the translation error (bottom row). Our DARE method shows improved results with respect to rotation and translation error compared to the baseline JRMPC, even when using empirically optimal re-sampling settings. }
	\label{fig:all}
\end{figure}

Complementary to the results on the combined VPS and TLS ETH dataset, we present recall plots for each dataset separately. We provide results for the facade dataset in TLS ETH, office dataset in TLS ETH, VPS indoor dataset, and VPS outdoor dataset. Our approach (DARE) is compared with the following methods: JRMPC \cite{evangelidis2014generative}, ICP \cite{ICP_PAMI92} and CPD \cite{myronenko2010point}. The recall curves for the separated datasets are collected in figure~\ref{fig:all}. 

\begin{figure}
	\centering
	\subfloat[Initital point sets\label{fig:fac_ex_in}]{\includegraphics[width=0.3\columnwidth]{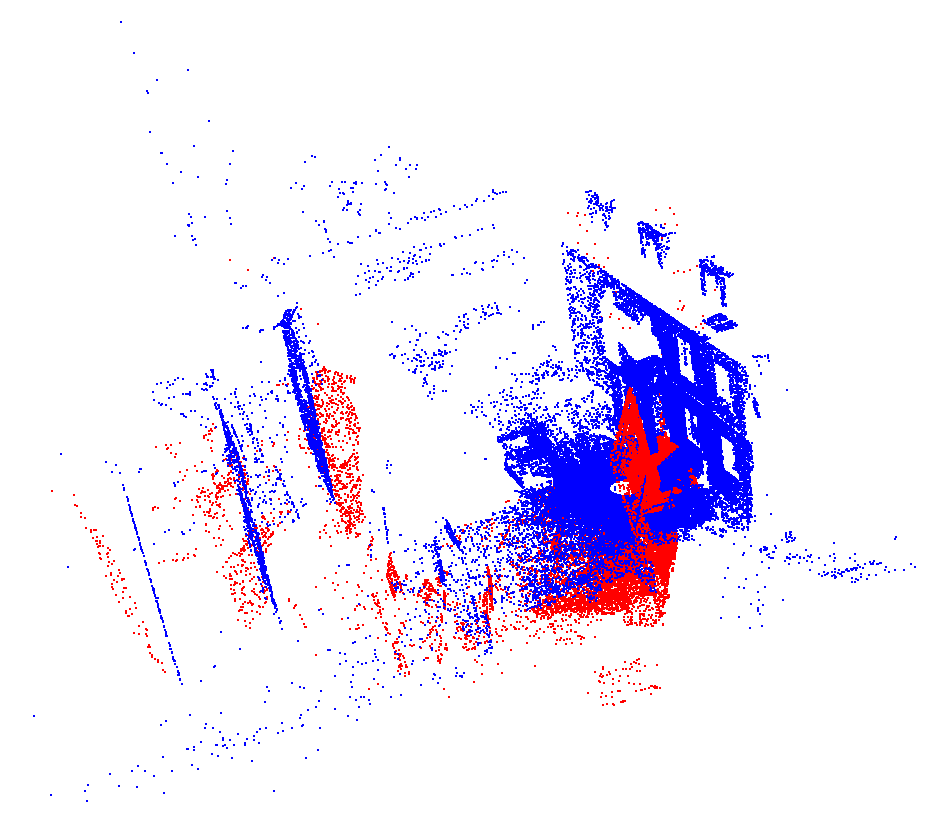}}
	\subfloat[Failed JRMPC registration\label{fig:fac_ex_jrmpc}]{\includegraphics[width=0.3\columnwidth]{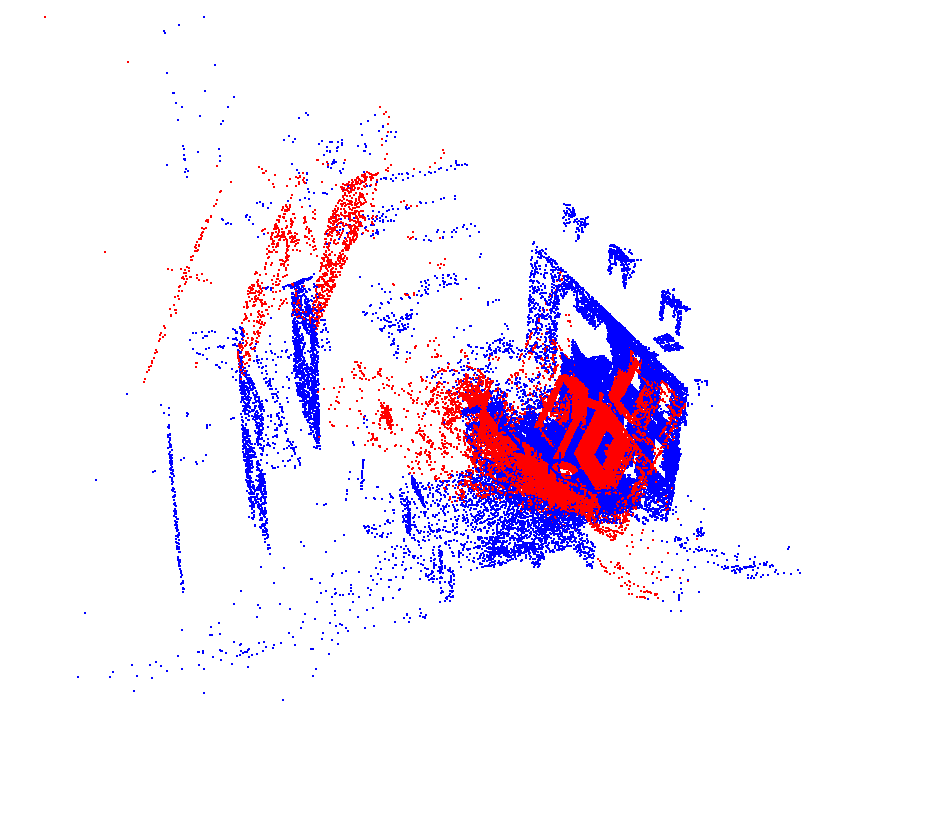}}
	\subfloat[Successful DARE registration\label{fig:fac_ex_ours}]{\includegraphics[width=0.3\columnwidth]{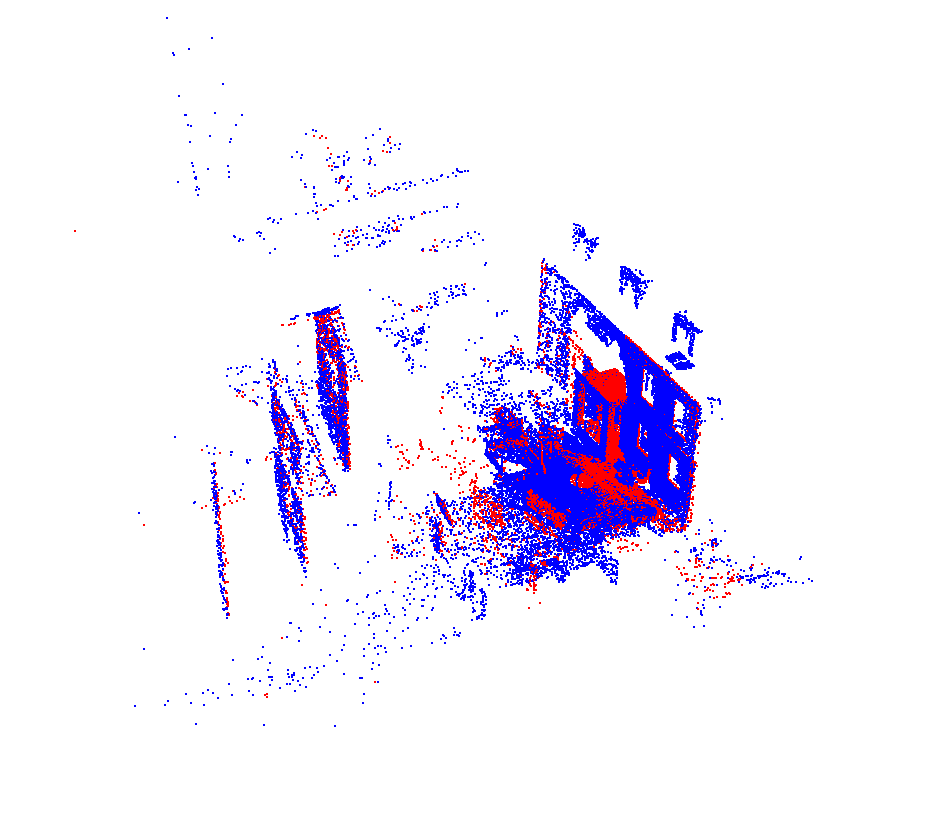}}
	\caption{Pairwise registration example on the facade dataset in TLS ETH. To distinguish between the different point sets they are colored in red and blue. Despite the large initial transformation error, seen in (a), our DARE method (c) successfully registers the two point sets.}\vspace{-2mm}
	\label{fig:facqual}
\end{figure}

\begin{figure}
	\centering
	\subfloat[Initiat point sets\label{fig:vps_fail_in}]{\includegraphics[width=0.3\columnwidth]{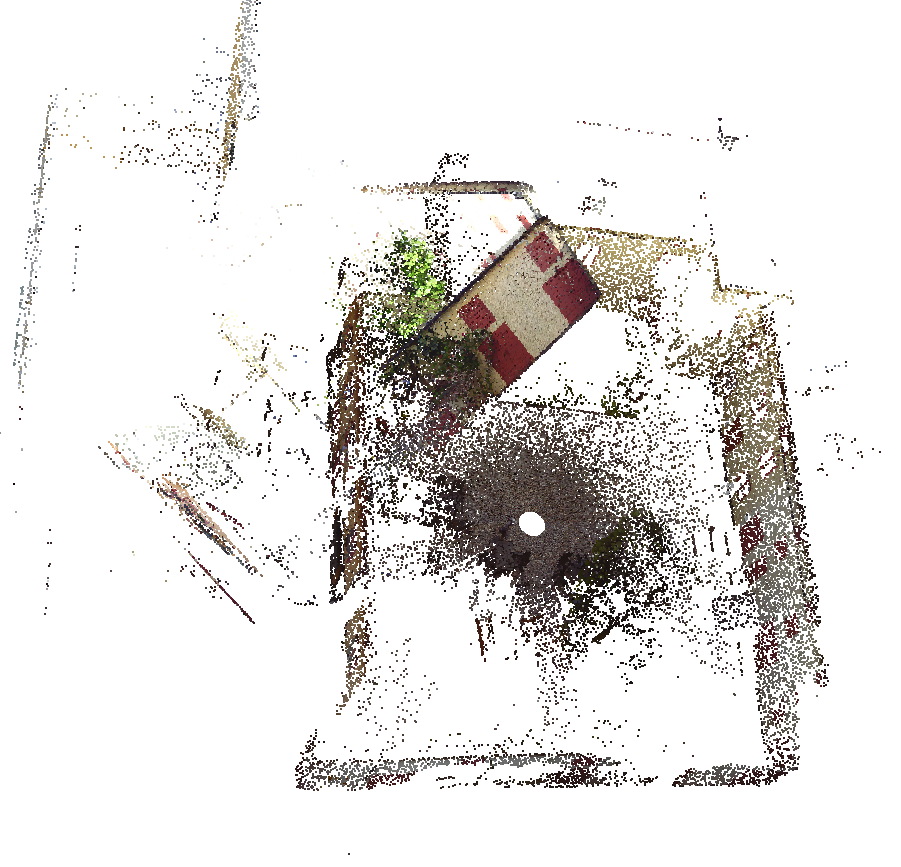}}
	\subfloat[Failed JRMPC registration\label{fig:vps_fail_jrmpc}]{\includegraphics[width=0.3\columnwidth]{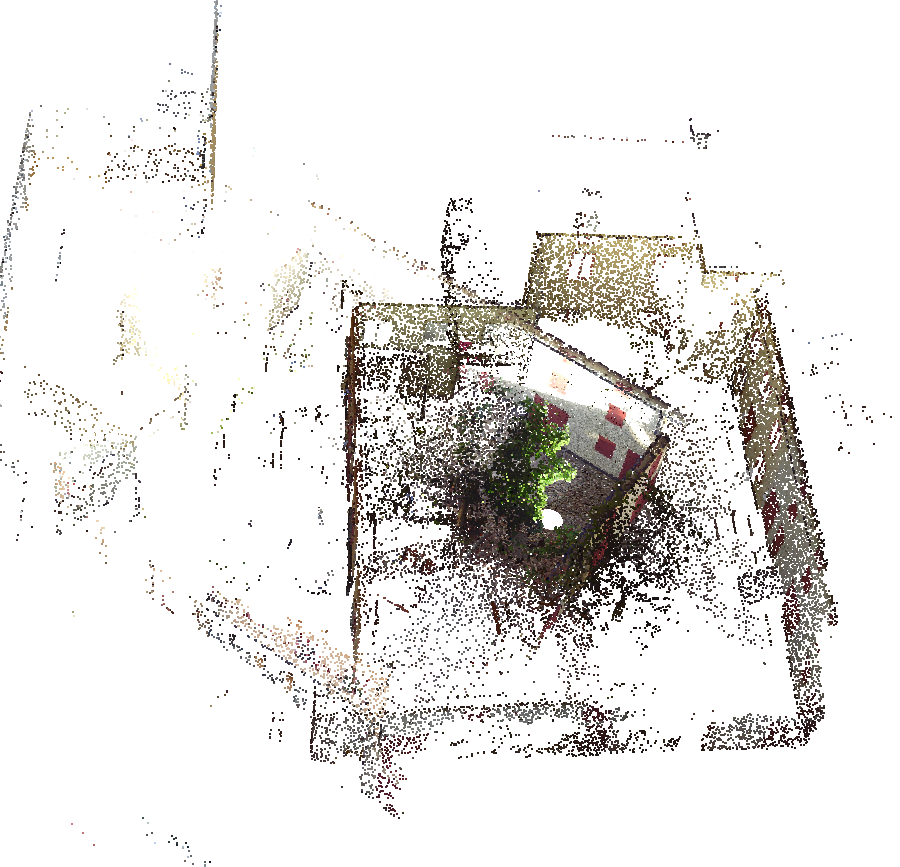}}
	\subfloat[Failed DARE registration\label{fig:vps_fail_ours}]{\includegraphics[width=0.3\columnwidth]{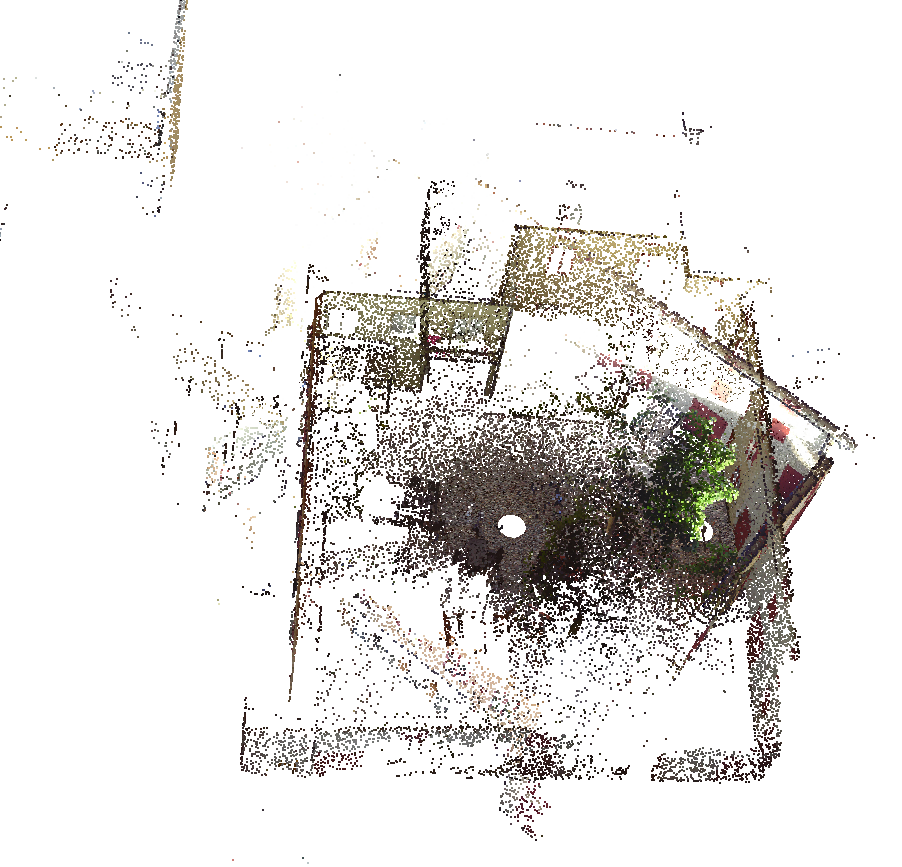}}
	\caption{Pairwise registration example on the indoor dataset in VPS with very limited overlap and large initial transformation error. Both the baseline JRMPC and our approach fails to align the point sets.}\vspace{-4mm}
	\label{fig:failedreg}
\end{figure}

Figure~\ref{fig:facqual} shows a qualitative comparison of our
approach, when performing pairwise registration, on the facade dataset
in TLS ETH. The baseline JRMPC method fails to register the point
clouds. Our approach successfully performs the registration task on
this dataset. We also provide a situation where our approach fails to
align the point sets. Figure~\ref{fig:failedreg} shows a pairwise
registration example on the VPS outdoor dataset. In this example, the
point sets had a very limited overlap due to the placement of the
Lidar sensor during acquisition. As we can see in figure \ref{fig:failedreg}, both the baseline JRMPC and our approach struggle, since none of the methods is designed to handle these extreme cases.

\begin{figure}
	\centering
	\includegraphics[width=0.4\columnwidth]{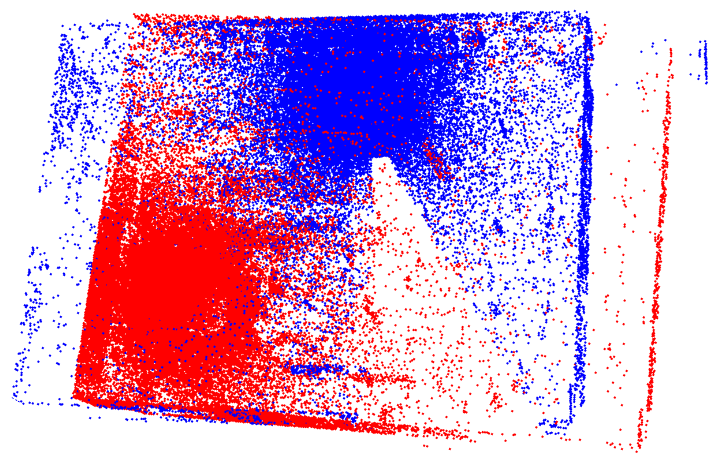}
	\caption{Failure example for our DARE method on the office dataset. The registration procedure has converged in a local minimum, where two of the aligned walls in the office room are are shifted, leading to a large translation error.}
	\label{fig:failtrans}
\end{figure}

Another failure case for our DARE method is shown in figure \ref{fig:failtrans}. In this example the registration procedure has converged in a local minimum, where two of the aligned walls in the office room are shifted. This partly explains the reduced recall with respect to the translation error in comparison to the recall with respect to the rotation error, seen in figure~ \ref{fig:full_cr}.

\subsection{Multi-view registration}

Finally, we present results, in case of multi-view registration, on the VPS indoor dataset. We compare our DARE approach with JRMPC and our color based approach DARE-color to CPPSR \cite{DanelljanCVPR2016}. We also compare our approaches to JRMPC and CPPSR with the optimal re-sampling settings from \ref{sec:rs}. 

\begin{figure}[!t]
	\centering
	\subfloat[Rotation recall\label{fig:vpsinjoint}]{\includegraphics[width=0.4\columnwidth]{totalRecall_jointvpsin_cr.pdf}}\hspace{2mm}
	\subfloat[Translation recall\label{fig:vpsinjoint_t}]{\includegraphics[width=0.4\columnwidth]{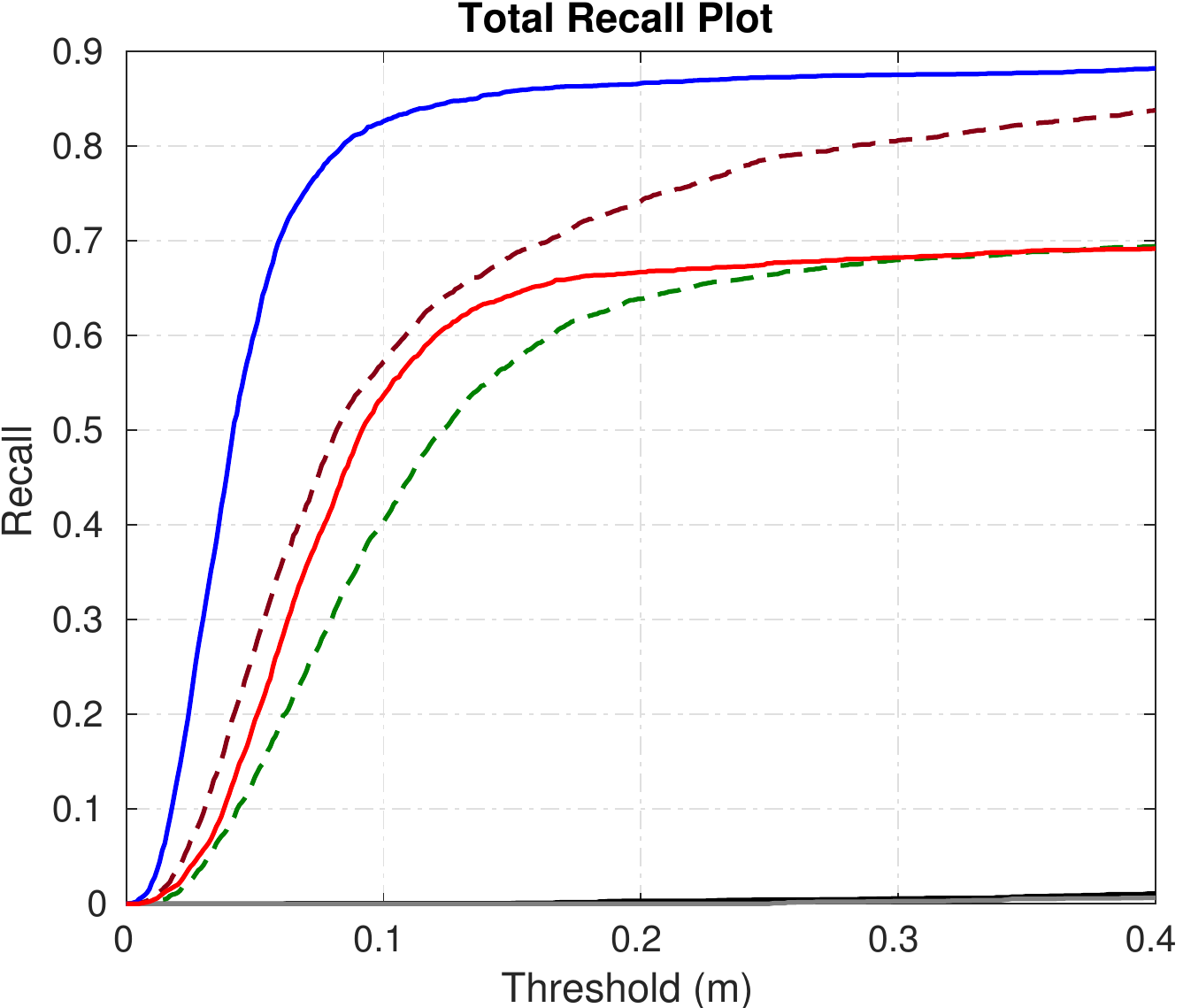}}
	\caption{Recall curves with respect to (a) the angular error and (b) translation error on the indoor dataset in VPS. Our DARE and DARE-color (DARE with color) approaches show significant improvement compare to both JRMPC and CPPSR. Our approach also shows improved results compared to optimal re-sampling.}\vspace{-2mm}
	\label{fig:multi_view_vps}
\end{figure}

\begin{figure}
	\centering
	\subfloat[Rotation recall\label{fig:vpsoutjoint}]{\includegraphics[width=0.4\columnwidth]{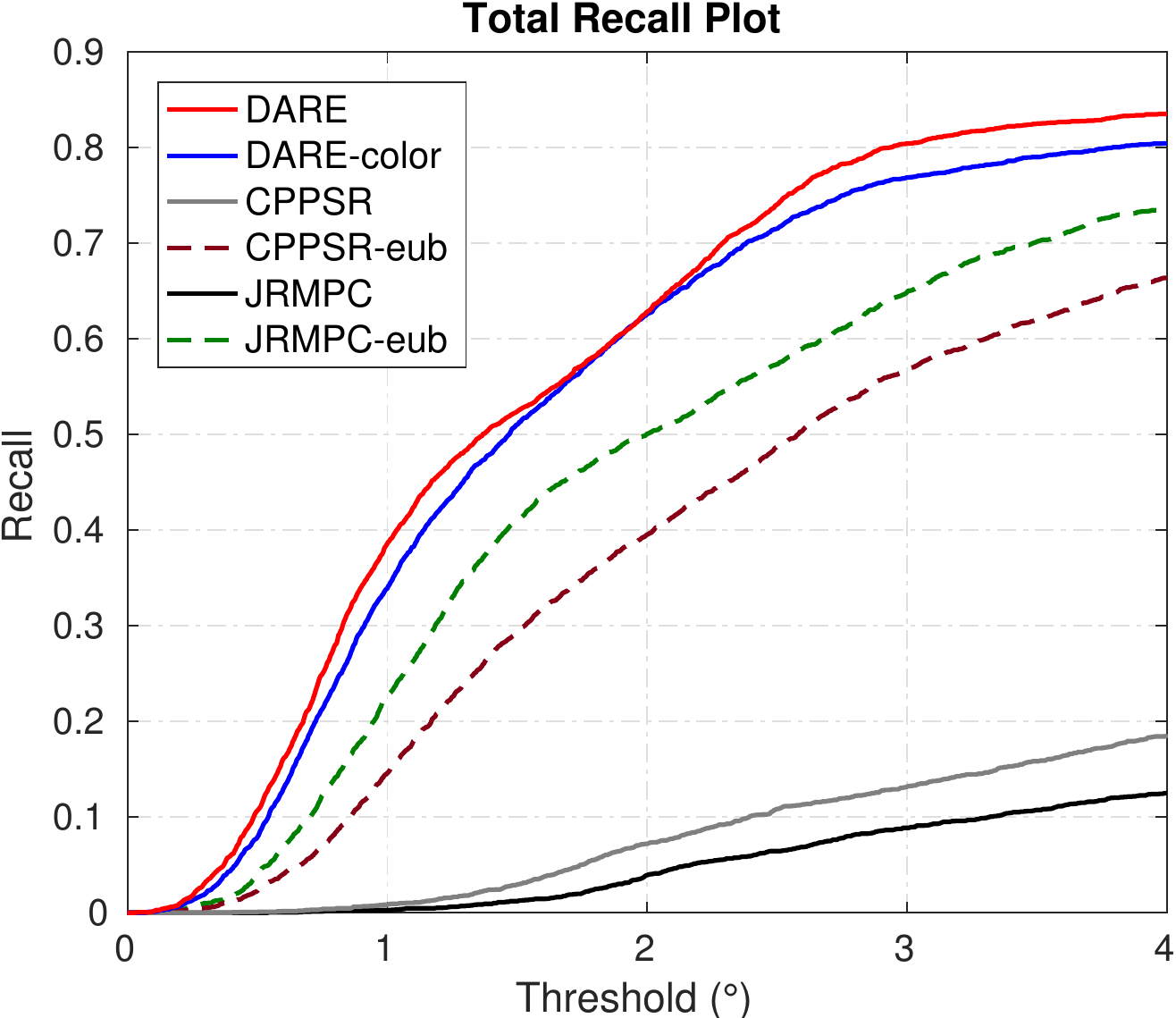}}\hspace{2mm}
	\subfloat[Translation recall\label{fig:vpsoutjoint_t}]{\includegraphics[width=0.4\columnwidth]{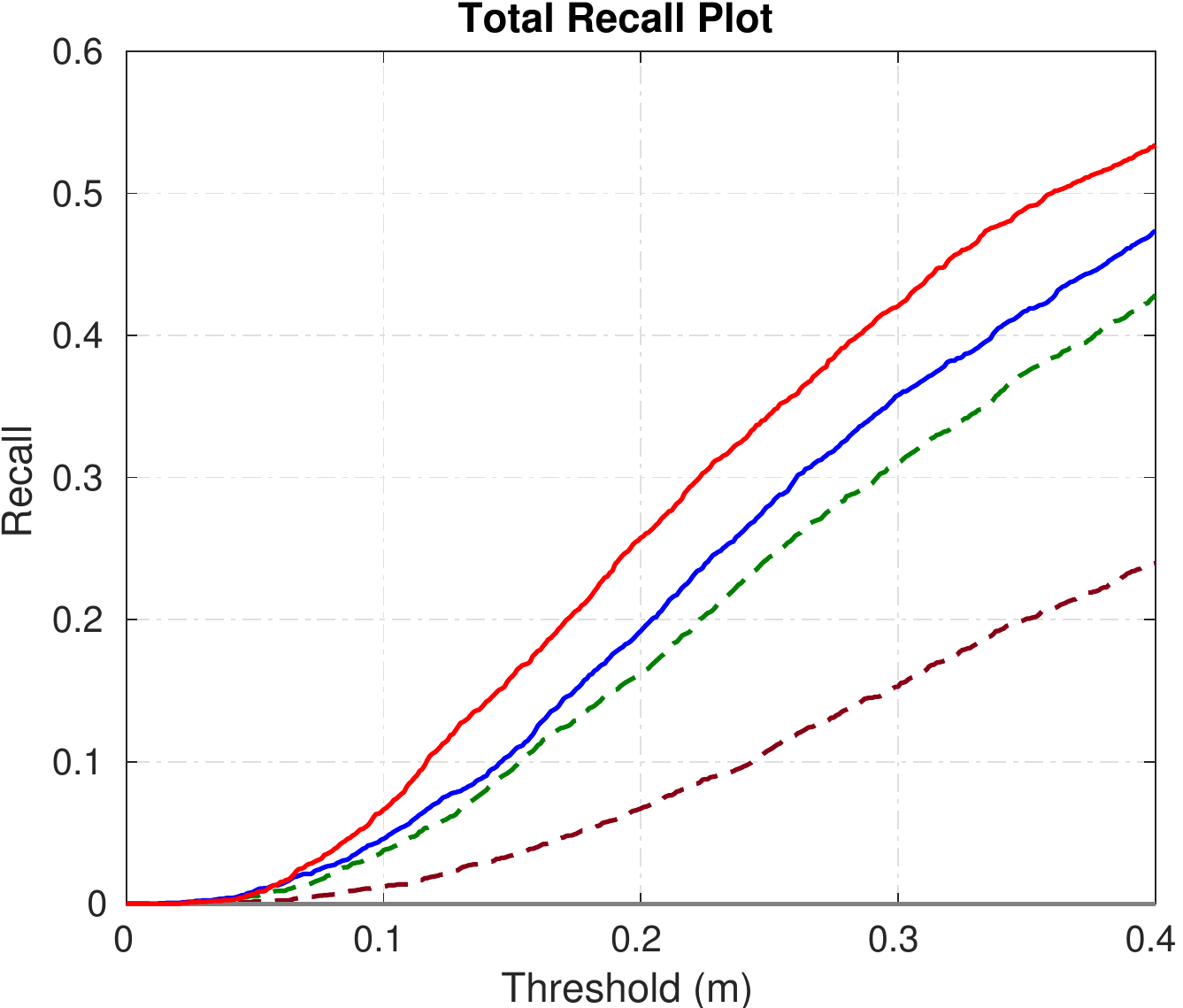}}
	\caption{Recall curves with respect to (a) the angular error and (b) translation error on the indoor dataset in VPS. Our DARE and DARE-color (DARE with color) approachess show significant improvement compare to both JRMPC and CPPSR. Our approach also shows improved results compared to optimal re-sampling.}\vspace{-2mm}
	\label{fig:multi_view_vpsout}
\end{figure}

The evaluation is performed over 500 registrations. The ground-truth is generated by first selecting a random rotation axis for each point set. We then rotate the point sets with rotation angles within 0-45 degrees around the rotation axes and apply random translations drawn from a multivariate normal distribution with standard deviation 1.0 in all directions. Figures ~\ref{fig:multi_view_vps} and ~\ref{fig:multi_view_vpsout} shows recall curves on the VPS indoor and outdoor datasets respectively. Our DARE method provides improved registration results compared to both CPPSR and JRMPC.

\end{document}